%% file: main.tex
\RequirePackage{snapshot}
\PassOptionsToPackage{numbers}{natbib}
\documentclass{article}
\usepackage[final]{neurips_2020}
\usepackage[utf8]{inputenc}
\usepackage[T1]{fontenc}

\usepackage{adjustbox}
\usepackage{amsfonts}
\usepackage{amsmath}
\usepackage{amssymb}
\usepackage{bm}
\usepackage{booktabs}
\usepackage{booktabs}
\usepackage{caption}
\usepackage{cuted}
\usepackage{float}
\usepackage{graphicx}
\usepackage{hyperref}
\usepackage{microtype}
\usepackage{nicefrac}
\usepackage{pbox}
\usepackage{setspace}
\usepackage{siunitx}
\usepackage{stmaryrd}
\usepackage{subfigure}
\usepackage{textcomp}
\usepackage{url}
\usepackage{varwidth}
\usepackage{xspace}
\usepackage{xcolor}

\usepackage{hyperref}
\usepackage{cleveref}

\newcommand{\ie}{\emph{i.e.}}

\IfFileExists{/Users/vedaldi/.bash_profile}{
\hbadness=\maxdimen
\vbadness=\maxdimen
\vfuzz=30pt
\hfuzz=30pt
}{}

\crefname{figure}{Figure}{Figures}
\Crefname{figure}{Figure}{Figures}
\crefname{table}{Table}{Tables}
\Crefname{table}{Table}{Tables}
\crefname{section}{Section}{Sections}
\Crefname{section}{Section}{Sections}

\makeatletter
\renewcommand{\paragraph}{%
  \@startsection{paragraph}{4}%
  {\z@}{0.10em}{-1em}%
  {\normalfont\normalsize\bfseries}%
}
\makeatother

\newcommand{\vect}[1]{\boldsymbol{\mathbf{#1}}}

\newcommand{\dpcl}{$d_\text{pcl}$\xspace}
\newcommand{\ddepth}{$d_\text{depth}$\xspace}
\newcommand{\overbar}[1]{\mkern 1.5mu\overline{\mkern-1.5mu#1\mkern-1.5mu}\mkern 1.5mu}

\title{Canonical 3D Deformer Maps:\\
Unifying parametric and non-parametric methods for \\dense weakly-supervised category reconstruction}

\author{David Novotny$^\ast$
\qquad
Roman Shapovalov\thanks{Authors contributed equally.}
\qquad
Andrea Vedaldi\\
Facebook AI Research\\
{\tt\small \{dnovotny, romansh, vedaldi\}@fb.com}\\
\small\url{http://www.robots.ox.ac.uk/\~david/c3dm/}}

\newcommand{\method}{C3DM\xspace}

\begin{document}
\maketitle
\begin{abstract}
We propose the \textit{Canonical 3D Deformer Map}, a new representation of the 3D shape of common object categories that can be learned from a collection of 2D images of independent objects.
Our method builds in a novel way on concepts from parametric deformation models, non-parametric 3D reconstruction, and canonical embeddings, combining their individual advantages.
In particular, it learns to associate each image pixel with a deformation model of the corresponding 3D object point which is canonical, i.e.~intrinsic to the identity of the point and shared across objects of the category.
The result is a method that, given only sparse 2D supervision at training time, can, at test time, reconstruct the 3D shape and texture of objects from single views, while establishing meaningful dense correspondences between object instances.
It also achieves state-of-the-art results in dense 3D reconstruction on public in-the-wild datasets of faces, cars, and birds.
\end{abstract}

\section{Introduction}\label{s:intro}

We address the problem of learning to reconstruct 3D objects from individual 2D images.
While 3D~reconstruction has been studied extensively since the beginning of computer vision research~\cite{Roberts1963}, and despite exciting progress in monocular reconstruction for objects such as humans, a solution to the general problem is still elusive.
A key challenge is to develop a \emph{representation} that can learn the 3D shapes of common objects such as cars, birds and humans from 2D images, without access to 3D ground truth, which is difficult to obtain in general.
In order to do so, it is not enough to model individual 3D shapes;
instead, the representation must also \emph{relate the different shapes} obtained when the object deforms (e.g.~due to articulation) or when different objects of the same type are considered (e.g.~different birds).
This requires establishing \emph{dense correspondences} between different shapes, thus identifying equivalent points (e.g.~the left eye in two birds).
Only by doing so, in fact, the problem of reconstructing independent 3D shapes from 2D images, which is ill-posed, reduces to learning a single deformable shape, which is difficult but approachable.

\input{fig-teaser}

In this paper, we introduce the \emph{Canonical 3D Deformer Map}~(\method), a representation that meets these requirements~(\cref{f:teaser}).
\method combines the benefits of parametric and non-parametric representations of 3D objects.
Conceptually, \method starts from a \emph{parametric 3D shape model} of the object, as often used in Non-Rigid Structure From Motion (NR-SFM~\cite{bregler2000recovering}).
It usually takes the form of a \emph{mesh} with 3D vertices $\vect{X}_1,\dots,\vect{X}_K\in\mathbb{R}^3$ expressed as a linear function of global deformation parameters~$\vect{\alpha}$, such that $\vect{X}_k = B_k \vect{\alpha}$ for a fixed operator~$B_k$.
Correspondences between shapes are captured by the identities~$k$ of the vertices, which are invariant to deformations.
Recent works such as Category-specific Mesh Reconstruction~(CMR)~\cite{kanazawa18learning} put this approach on deep-learning rails, learning to map an image~$I$ to the deformation parameters~$\vect{\alpha}(I)$.
However, working with meshes causes a few significant challenges, including guaranteeing that the mesh does not fold, rendering the mesh onto the image for learning, and dealing with the finite mesh resolution.
It is interesting to compare parametric approaches such as CMR to \emph{non-parametric depth estimation models}, which directly map each pixel~$\vect{y}$ to a depth value~$d_{\vect{y}}(I)$~\cite{Yao2018,Khot2019,godard17unsupervised}, describing the geometry of the scene in a dense manner.
The depth estimator $d_{\vect{y}}(I)$ is easily implemented by means of a convolutional neural network and is not bound to a fixed mesh resolution.
However, a depth estimator has no notion of correspondences and thus of object deformations.

Our intuition is that these two ways of representing geometry, parametric and non-parametric, can be combined by making use of the third notion, a \emph{canonical map}~\cite{thewlis17bunsupervised,schmidt17self-supervised,kulkarni19canonical}.
A canonical map is a non-parametric model~$\Phi_{\vect{y}}(I) = \kappa$ that associates each pixel~$\vect{y}$ to the intrinsic coordinates~$\kappa$ of the corresponding object point.
The latter can be thought of as a continuous generalization of the index~$k$ that in parametric models identifies a vertex of a mesh.
Our insight is that \emph{any} intrinsic quantity --- i.e.~one that depends only on the identity of the object point --- can then be written as a function of~$\kappa$.
This includes the \emph{3D deformation operator}~$B_\kappa$, so that we can reconstruct the 3D point found at pixel~$\vect{y}$ as $\vect{X}_{\vect{y}}= B_\kappa \vect{\alpha}$.
Note that this also requires to learn the mapping~$\kappa \mapsto B_\kappa$, which we can do by means of a small neural network.

We show that the resulting representation, \method, can reconstruct the shape of 3D objects densely and from single images, using only easily-obtainable 2D supervision at training time  --- the latter being particularly useful for 3D reconstruction from traditional non-video datasets.
We extensively evaluate \method and compare it to CMR~\cite{kanazawa18learning}, state-of-the-art method for monocular category reconstruction.
\method achieves both higher 3D reconstruction accuracy and more realistic visual reconstruction on real-world datasets of birds, human faces, and four other deformable categories of rigid objects.

\section{Related work}\label{s:related}

The literature contains many impressive results on image-based 3D reconstruction.
To appreciate our contribution, it is essential to characterize the assumptions behind each method, the input they require for training, and the output they produce.
Multiple works~\cite{loper15smpl,anguelov05scape,guan09estimating,sigal08combined,bogo16keep,lassner17unite,huang17towards,zanfir18monocular,joo18total,pavlakos19expressive,xiang19monocular,tan17indirect,tung17self-supervised,omran18neural,pavlakos18learning,kanazawa18end-to-end,kolotouros19convolutional,sigal08combined,lassner17unite,rogez18lcr-net,pavlakos18learning,varol18bodynet,pavlakos19expressive}
take as input an existing parametric 3D model of the deformable object such as SMPL~\cite{loper15smpl} or SCAPE~\cite{anguelov05scape} for humans bodies, or Basel~\cite{bfm09} for faces and \emph{fit it to images}.
In our case, no prior parametric 3D model is available; instead, our algorithm \emph{simultaneously learns and fits a 3D model using only 2D data as input}.

\paragraph{Sparse NR-SFM\!\!\!\!}
methods receive sparse 2D keypoints as input and lift them in 3D, whereas \method receives as input an image and produces a \textit{dense} reconstruction.
In other words, we wish to obtain dense reconstruction of the objects although only sparse 2D annotations are still provided during training.
For learning, NR-SFM methods need to separate the effect of viewpoint changes and deformations~\cite{xiao2004dense}.
They acheive it by constraining the space of deformations in one of the following ways:
assume that shapes span a low-rank subspace~\cite{agudo2018image,fragkiadaki2014grouping,dai2014simple,zhu2014complex}
or that 3D trajectories are smooth in time~\cite{akhter2009nonrigid, akhter2011trajectory}, or combine both types of constraints~\cite{agudo2017dust,gotardo2011non, kumar2018scalable,kumar2017spatial},
or use multiple subspaces~\cite{zhu2014complex,agudo2018deformable}, sparsity~\cite{zhou2016sparseness,zhou2016sparse} or Gaussian priors~\cite{torresani2008nonrigid}.
In \Cref{s:learning}, we use NR-SFM to define one of the loss functions.
We chose to use the recent C3DPO method~\cite{novotny19c3dpo}, which achieves that separation by training a canonicalization network, due to its state-of-the-art performance.

\paragraph{Dense 3D reconstruction.}

Differently from our work, most of the existing approaches to dense 3D reconstruction assume either 3D supervision or rigid objects and multiple views.
Traditional \emph{multi-view} approaches~\cite{ayache86efficient} perform 3D reconstruction by analyzing disparities between two or more calibrated views of a rigid object
(or a non-rigid object simultaneously captured by multiple cameras), but may fail to reconstruct texture-less image regions.
Learning multi-view depth estimators with~\cite{Yao2018} or without~\cite{Khot2019} depth supervision can compensate for lacking visual evidence.
The method of~\citet{Innmann2019} can reconstruct mildly non-rigid objects, but still requires multiple views.

Most methods for \emph{single-view} dense reconstruction of object categories require 3D supervision~\cite{Mescheder2019,Tatarchenko2019,Groueix2018}.
In particular, AtlasNet~\cite{Groueix2018} uses a representation similar to ours, mapping points on a two-dimensional manifold to points on the object's surface with a multi-layer perceptron~(MLP).
Instead of conditioning the MLP on a shape code, we map the manifold points to embeddings; the 3D location is defined as their linear combination.
Only a few methods, like \method, manage to learn parametric shape from 2D data only: \citet{cashman2013shape} propose a morphable model of dolphins supervised with 2D keypoints and segmentation masks, while others~\cite{vicente2014reconstructing,carreira2015virtual} reconstruct the categories of PASCAL VOC\@.
Most of these methods start by running a traditional SFM pipeline to obtain the mean 3D reconstruction and camera matrices.
\citet{kar2015category} replace it with NR-SFM for reconstructing categories of PASCAL3D+.
VpDR~\cite{novotny17learning} reconstructs \emph{rigid} categories from monocular views.
\citet{Wu2020} reconstruct non-rigid symmetrical shapes by rendering predicted depth maps, albedo, and shading and ensuring symmetry, which works well for limited viewpoint variation.
In this work, \method reconstructs 3D shape from a single image without assuming symmetry, limited range of viewpoints, or images related by rigid transform at training or prediction time.

A number of recent methods based on \emph{differentiable mesh rendering} can also be trained with 2D supervision only.
\citet{kanazawa18learning} introduced CMR, a deep network that reconstructs shape and texture of deformable objects;
it is the closest to our work in terms of assumptions, type of supervision, and output, and is currently state of the art for reconstruction of categories other than humans.
\mbox{DIB-R}~\cite{Chen2019} improves the rendering technique by softly assigning all image pixels, including background, to the mesh faces.
Working with meshes is challenging since the model should learn to generate only valid meshes, e.g.~those without face intersections.
\citet{Henderson2020} proposed parametrisation of a mesh that prevents intersecting faces.
In contrast to these methods, we work with point clouds and avoid computationally expensive rendering by leveraging NR-SFM pre-processing and cross-image consistency constraints.
The concurrent work, Implicit Mesh Reconstruction~\cite{Tulsiani2020}, defines similar constraints to our reprojection and cross-image consistency
using rendering-like interpolation of 3D points on the mesh surface.
We avoid this step by predicting 3D coordinates of each pixel in a feed-forward manner.
IMR does not build, as we do, on NR-SFM.
The advantage is that this enables a variant that trains without keypoint supervision.
The disadvantage is that, in order do to so, IMR has to initialise the model with a hand-crafted category-specific template mesh.

\paragraph{Canonical maps.}

A \emph{canonical map} is a function that maps image pixels to identifiers of the corresponding object points.
Examples include the UV surface coordinates used by Dense Pose~\cite{guler18densepose:} and spherical coordinates~\cite{thewlis17bunsupervised}.
\citet{thewlis17bunsupervised,thewlis19unsupervised,schmidt17self-supervised} learn canonical maps in an unsupervised manner via a bottleneck, whereas~\citet{kulkarni19canonical,Kulkarni2020} do so by using consistency with an initial 3D model.
Normalized Object Coordinate Space (NOCS)~\cite{Wang2019} also ties canonical coordinates and object pose, however it does not allow for shape deformation;
different shapes within category have to be modelled by matching to one of the hand-crafted exemplars.
Instead, we learn the dense parametric 3D deformation model for each object category from 2D data.



\section{Canonical 3D Deformer Map representation}\label{s:method}

\begin{figure*}
\centering\includegraphics[width=\textwidth]{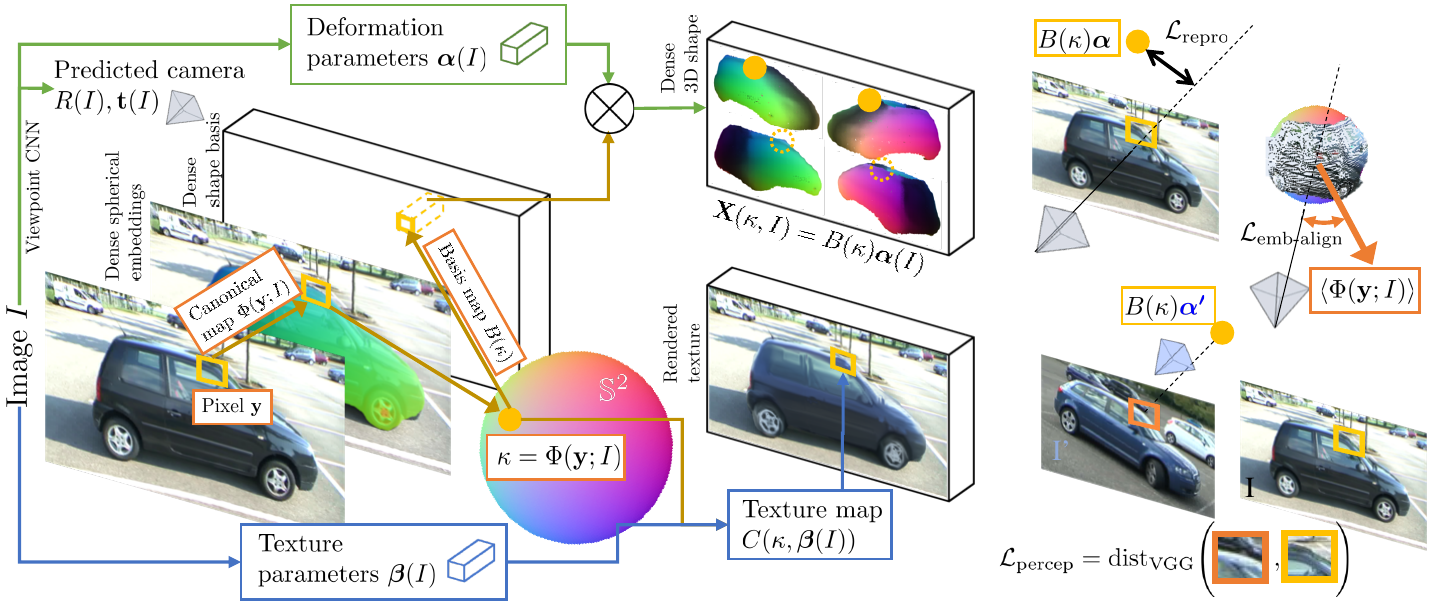}
\caption{
\textbf{Detailed system overview.}
At test time, the image is passed through the network~$\Phi$ to obtain the map of dense embeddings~$\kappa \in \mathbb{S}^2$.
The network~$B$ converts them individually to deformation operators.
In the meantime, the image is passed to the viewpoint network to predict the camera orientation~$R$ and shape parameters~$\vect{\alpha}$.
Eq.\,\eqref{e:X} combines these quantities to obtain 3D reconstruction for each pixel within the object mask.
At training time, sparse 2D keypoints are preprocessed with C3DPO~\cite{novotny19c3dpo} to obtain ``ground truth'' camera orientation~$R^*$ and shape parameters~$\vect{\alpha}^*$.
These, together with the C3DPO basis~$B^*$, are used in~\eqref{e:basis_loss} to supervise the corresponding predicted variables.
On the right, three more loss functions are illustrated: reprojection loss~\eqref{e:repro_loss}, cross-projection perceptual loss~\eqref{e:crossproj}, and~\eqref{e:emb-align} aligning the camera orientation with average embedding direction.
}\label{f:overview}
\end{figure*}


\subsection{The model}\label{s:representation}

\paragraph{Canonical map.}

Let $I \in \mathbb{R}^{3\times H\times W}$ be an image and~$\Omega \subset \{1,\dots,H\}\times \{1,\dots,W\}$ be the image region that contains the object of interest.
We consider a \emph{canonical map}~$\kappa = \Phi(\vect{y};I)$ sending pixels $\vect{y}\in\Omega$ to points on the unit sphere $\kappa\in\mathbb{S}^2$,
which is topologically equivalent to any 3D surface $\mathcal{S}\subset\mathbb{R}^3$ without holes.
It can be interpreted as a space of indices or coordinates $\kappa$ that identify a dense system of `landmarks' for the deformable object category.
A landmark, such as the corner of the left eye in a human, is a point that can be identified repeatably despite object deformations.
Note that the index space can take other forms than~$\mathbb{S}^2$, however the latter is homeomorphic to most surfaces of 3D objects and has the minimum dimensionality, which makes it a handy choice in practice.

\paragraph{Deformation model.}

We express the 3D location of a landmark~$\kappa$ as~$\vect{X}(\kappa;I) = B(\kappa) \vect{\alpha}(I)$, where $\alpha(I)\in\mathbb{R}^D$ are image-dependent deformation parameters and~$B(\kappa) \in \mathbb{R}^{3\times D}$ is a linear operator indexed by $\kappa$.
This makes~$B(\kappa)$ an \emph{intrinsic} property, invariant to the object deformation or viewpoint change.
The full 3D reconstruction~$\mathcal{S}$ is given by the image of this map:
$
\mathcal{S}(I) = \{ B(\kappa) \vect{\alpha}(I): \kappa \in \mathbb{S}^2 \}.
$
The reconstruction~$\vect{X}(\vect{y};I)$ specific to the pixel~$\vect{y}$ is instead given by composition with the canonical map:
\begin{equation}\label{e:X}
  \vect{X}(\vect{y};I) = B(\kappa) \vect{\alpha}(I),
  \quad \text{where}~~~
  \kappa = \Phi(\vect{y};I).
\end{equation}

\paragraph{Viewpoint.}

As done in NR-SFM, we assume that the 3D reconstruction is `viewpoint-free', meaning that the viewpoint is modelled not as part of the deformation parameters~$\vect{\alpha}(I)$, but explicitly, as a separate rigid motion $(R(I),\vect{t}(I))\in \mathbb{SE}(3)$.
The rotation~$R$ is regressed from the input image in the form proposed by \citet{Zhou2019}, and translation~$\vect{t}(I)$ is found by minimizing the reprojection, see \cref{s:learning} for details.
We assume to know the perspective/ortographic \emph{camera model} $\pi : \mathbb{R}^3 \rightarrow \mathbb{R}^2$ mapping 3D points in the coordinate frame of the camera to 2D image points (see sup.~mat.~for details).
With this, we can recover the coordinates~$\vect{y}$ of a pixel from its 3D reconstruction~$\vect{X}(\vect{y};I)$ as:
\begin{equation}\label{e:y}
  \vect{y} = \pi\left(
  R(I) \vect{X}(\vect{y};I) + \vect{t}(I)
  \right).
\end{equation}
Note that~$\vect{y}$ appears on both sides of \cref{e:y};
this lets us define the self-consistency constraint~\eqref{e:repro_loss}.

\paragraph{Texture.}

In addition to the deformation operator $B(\kappa)$, any intrinsic property can be descried in a similar manner.
An important example is reconstructing the \emph{albedo} $I(\vect{y})$ of the object, which we model as:
\begin{equation}\label{e:tex-model}
   I(\vect{y}) = C(\kappa;\vect{\beta}(I)),
   ~~~
   \kappa = \Phi(\vect{y};I),
\end{equation}
where $C(\kappa;\vect{\beta})$ maps a small number of image-specific texture parameters $\vect{\beta}(I)\in\mathbb{R}^{D'}$ to the color of landmark~$\kappa$.
In \cref{s:result-corresp}, we use this model to transfer texture between images of different objects.

\paragraph{Implementation via neural networks.}

The model above includes several learnable functions that are implemented as deep neural networks.
In particular, the canonical map~$\Phi(I)$ is implemented as an image-to-image convolutional network (CNN) with an~$\mathbb{R}^{3\times H\times W}$ input (a color image) and an~$\mathbb{R}^{3\times H\times W}$ output (the spherical embedding).
The last layer of this network normalizes each location in $\ell^2$ norm to project 3D vectors to~$\mathbb{S}^2$.
Functions $\vect{\alpha}(I)$, $\vect{\beta}(I)$ and $R(I)$ predicting deformation, texture and viewpoint rotation are also implemented as CNNs.
Translation~$\vect{t}$ is found by minimising the reprojection, as explained below.
Finally, functions~$B(\kappa)$ and~$C(\kappa)$ mapping embeddings to their 3D deformation and texture models are given by multi-layer perceptrons (MLP).
The latter effectively allows~$\kappa$, and the resulting 3D and texture reconstruction, to have arbitrary resolution.

\subsection{Learning formulation}\label{s:learning}

The forward pass and loss functions are shown in~\Cref{f:overview}. In order to train \method, we assume available a collection of independently-sampled views of an object category $\{I_n\}_{n=1}^N$.%
\footnote{``Independent'' means that views contain different object deformations or even different object instances.}
Furthermore, for each view, we require annotations for the silhouette~$\Omega_n$ of the object as well as the 2D locations of~$K$ landmarks~$Y_n = (\vect{y}_{n1},\dots,\vect{y}_{nK})$.
In practice, this information can often be extracted automatically via a method such as Mask R-CNN~\cite{he2017mask} and HRNet~\cite{Sun2019}, which we do for most experiments in~\cref{s:exp}.
Note that \method requires only a small finite set of $K$~landmarks for training, while it learns to produce a continuous landmark map.
We use the deformation basis from an NR-SFM method as a prior and add a number of consistency constraints for self-supervision, as discussed next.

\paragraph{NR-SFM Prior.}

Since our model generalizes standard parametric approaches, we can use any such method to bootstrap and accelerate learning.
We use the output of the recent C3DPO~\cite{novotny19c3dpo} algorithm $\mathcal{A}^*_n = (B^*_n, \mathcal{V}^*_n, \vect{\alpha}^*_n, R^*_n)$ in order to anchor the deformation model~$B(\kappa)$ in a visible subset~$\mathcal{V}^*_n$ of~$K$ discrete landmarks,
as well as the deformation and viewpoint parameters, for each training image~$I_n$.

Note that, contrary to \method, C3DPO takes as input the 2D location of the sparse keypoints both at training \emph{and test time}.
Furthermore, it can only learn to lift the keypoints for which ground-truth is available at training time.
In order to learn \method, we thus need to learn from scratch the deformation and viewpoint networks~$\vect{\alpha}(I)$ and~$R(I)$, as well as the continuous deformation network~$B(\kappa)$.
This is necessary so that at test time \method can reconstruct the object in a dense manner given only the image~$I$, not the keypoints, as input.
At training time, we supervise the deformation and viewpoint networks from the C3DPO output via the loss:
\begin{equation}\label{e:basis_loss}
\mathcal{L}_\text{pr}(\Phi,B,\vect{\alpha},R;I,Y,\mathcal{A}^*)
=
\frac{1}{| \mathcal{V}^* |}
\sum_{k \in \mathcal{V}^*}
\| B(\Phi(\vect{y}_k;I))  - B^*_k \|_\epsilon
+ w_{\vect{\alpha}} \| \vect{\alpha}(I) - \vect{\alpha}^* \|_\epsilon
+ w_{R} d_\epsilon(R(I);R^*),
\end{equation}
where $\| z \|_\epsilon$ is the pseudo-Huber loss~\cite{Charbonnier97} with soft threshold\,$\epsilon$ and $d_\epsilon$ is a distance between rotations.%
\footnote{
  $\| z \|_\epsilon = \epsilon (\sqrt{1+(\|z\|/\epsilon)^2}-1)$; it behaves as a quadratic function of~$\| z \|$ in the vicinity of 0 and a linear one when $\| z \| \rightarrow \infty$, which makes it both smooth and robust to outliers.
See sup.\,mat.~for definition of~$d_\epsilon$.
}

\paragraph{Projection self-consistency loss.}

As noted in \cref{s:related}, the composition of~\cref{e:X,e:y} must yield the identity function.
This is captured by the \emph{reprojection consistency loss}
\begin{equation}\label{e:repro_loss}
\mathcal{L}_\text{repro}(\Phi, B, \vect{\alpha}, R; \Omega, I) =
\min_{\vect{t}\in\mathbb{R}^3}
\sum_{\vect{y}\in\Omega}
\left\|
\hat{\vect{y}}(\vect{t})
- \vect{y}
\right\|_\epsilon,
~~~
\hat{\vect{y}}(\vect{t}) =
\pi\Big(
  R(I)~ B(\Phi(\vect{y};I))~ \vect{\alpha}(I) + \vect{t}
\Big).
\end{equation}
It causes the 3D reconstruction of an image pixel~$\vect{y}$, which is obtained in a viewpoint-free space, to line up with~$\vect{y}$ once the viewpoint is accounted for.
We found optimizing over translation~$\vect{t}$ in~\cref{e:repro_loss} to obtain~$\vect{t}(I, \Omega, \Phi, B, \vect{\alpha}, R)$ based on the predicted shape to be more accurate than regressing it directly.
Refer to \Cref{s:details_camera} in sup.~mat.~for optimization algorithm.
We use the obtained value as the translation prediction~$\vect{t}(I)$, in particular, in \cref{e:crossproj}, only implying the dependency on the predictors to simplify the notation.
We backpropagate gradients from all losses through this minimization though.

\paragraph{Apperance loss.}

Given two views~$I$ and~$I'$ of an object,
we can use the predicted geometry and viewpoint to establish dense correspondences between them.
Namely, given a pixel~$\vect{y}\in\Omega$ in the first image, we can find the corresponding pixel~$\hat{\vect{y}}'$ in the second image as:
\begin{equation}\label{e:crossproj}
  \hat{\vect{y}}' =
  \pi\Big(
    R(I')~ B(\Phi(\vect{y};I))~ \vect{\alpha}(I') + \vect{t}(I')
  \Big).
\end{equation}
This equation is similar to~\cref{e:repro_loss}, in particular, the canonical map is still computed in the image~$I$ to identify the landmark, however the shape $\vect{\alpha}$ and viewpoint $(R,\vect{t})$ are computed from another image~$I'$.
Assuming that color constancy holds, we could then simply enforce~$I(\vect{y}) \approx I'(\hat{\vect{y}}')$, but this constraint is violated for non-Lambertian objects or images of different object instances.
We thus relax this constraint by using a \emph{perceptual loss} $\mathcal{L}_\text{percep}$, which is based on comparing the activations of a pre-trained neural network instead~\cite{Zhang2018}.
Please refer to \Cref{s:details_percept} in the sup.~mat.~for details.

Due to the robustness of the perceptual loss, most images~$I$ can be successfully matched to a fairly large set $\mathcal{P}_I = \{I'\}$ of other images, even if they contain a different instance.
To further increase robustness to occlusions and illumination differences caused by change of viewpoint, we follow~\citet{Khot2019}: given a batch of training images~$\mathcal{P}_I$, we compare each pixel in~$I$ only to $k \leq |\mathcal{P}_I|$ its counterparts in the batch that match the pixel best.
This bring us to the following formulation:
\begin{equation}\label{e:percep_k_loss}
\mathcal{L}^\text{min-k}_\text{percep}(\Phi,B,\vect{\alpha},R,\vect{t};\Omega,I,\mathcal{P}_I) =
\frac{1}{k}\sum_{\vect{y} \in \Omega}
\min_{Q \subset \mathcal{P}_I : |Q| = k}
\sum_{I'\in Q} \mathcal{L}_\text{percep}(\Phi,B,\vect{\alpha},R,\vect{t}; \vect{y},I,I').
\end{equation}

\paragraph{Learning the texture model.}
The texture model $(C,\beta)$ can be learned in a similar manner, by minimizing the combination of the photometric and perceptual~\eqref{e:percep_k_loss} losses between the generated and original image.
Please refer to the supplementary material for specific loss formulations.
We do not back-propagate their gradients beyond the appearance model as it deteriorates the geometry.

\paragraph{Camera-embedding alignment.}
We use another constraint that ties the spherical embedding space and camera orientation.
It forces the model to use the whole embedding space and avoid re-using its parts for the regions of similar appearance, such as left and right sides of a car.
We achieve it by aligning the direction of the mean embedding vector~$\kappa$ with the camera direction, minimizing
\begin{equation}\label{e:emb-align}
\mathcal{L}_\text{emb-align}(\Phi,R;\Omega,I)
=
\begin{bmatrix}
0 & 0 & 1
\end{bmatrix}
R(I) \frac{\overbar{\kappa}}{\|\overbar{\kappa}\|}, \quad \textrm{where}~~
\overbar{\kappa} = \frac{1}{|\Omega|}
\sum_{\vect{y}\in\Omega}
 \Phi(\vect{y};I).
\end{equation}

\paragraph{Mask reprojection loss.}
We observed that on some datasets like CUB Birds, the reconstructed surface tends to be noisy due to some parts of the embedding space overfitting to specific images.
To prevent it interfering with other images, we additionally minimize the following simple loss function:
\begin{equation}\label{e:mask}
\mathcal{L}_\text{mask}(B, \vect{\alpha}, R, \vect{t}; \Omega) =
\int_{\mathbb{S}^2} 
\bigg\llbracket \pi\Big(
  R~ B(\kappa)~ \vect{\alpha} + \vect{t}
\Big) \notin \Omega \bigg \rrbracket
d\kappa,
\end{equation}
where we approximate the integration by taking a uniform sample of 1000 points~$\kappa$ on a sphere.

\section{Experiments}\label{s:exp}

We evaluate the proposed method on several datasets using 3D reconstruction metrics.
It is difficult to quantitatively evaluate canonical maps, as discussed in \cref{s:result-corresp}.
Since reconstruction relies on having a good model for canonical mapping, we use that application to demonstrate the quality of produced maps.
For visual evaluation, we use another application, texture transfer, in \cref{f:tex-transfer}.

\paragraph{Implementation details.}

We build on the open-source implementation of C3DPO for pre-processing\footnote{\url{https://github.com/facebookresearch/c3dpo_nrsfm}} and set $\vect{\alpha}\in\mathbb{R}^{10}$, $\vect{\beta}\in\mathbb{R}^{128}$.
The canonical map network~$\Phi$ uses the Hypercolumns architecture~\cite{Hariharan2015} on top of ResNet-50~\cite{he2016deep}, while basis and texture networks~$B$ and~$C$ are MLPs.
See \Cref{s:arch_details,s:details_hyperp} in sup.~mat.~for description of the architecture, hyperparameters and optimization.

\input{tex-transfer-new}

\paragraph{Benchmarks.}

We evaluate the method on a range of challenging datasets.
We use \method to generate from each test image: (1) a full 360\textdegree~shape reconstruction as a point cloud~$\{B(\kappa)\vect{\alpha}(I): \kappa \in \mathcal{K}\}$, where~$\mathcal{K}$ consists of 30k sampled embeddings from random training set images,
and (2) a depth map from the estimated image viewpoint obtained for each pixel~$\vect{y} \in \Omega$ as the coordinate~$z$ of $R \vect{X}(\vect{y}; I)$.
We compare the full reconstructions against ground-truth point clouds using symmetric Chamfer distance~\dpcl (after ICP alignment~\cite{besl1992method}) and, whenever the dataset has depth maps or calibrations to project the ground-truth meshes, predicted depth maps against ground-truth depth maps as the average per-pixel depth error \ddepth.
In particular, to compute the symmetric Chamfer distance between the predicted and ground-truth point clouds~$d_{\textrm{pcl}}(\hat{C}, C)$, we first correct the scale ambuiguity by normalising the variance of the predicted point cloud to match ground truth.
Then, we align them with ICP to obtain the $\tilde{C} = sR\hat{C} + \vect{t}$ rigidly aligned with~$C$.
We define Chamfer distance as the mean $\ell^2$ distance from each point in~$C$ to its nearest neighbour in~$\tilde{C}$ and make it symmetric: 
\begin{equation}\label{e:mask}
d_{\textrm{pcl}}(\hat{C}, C) = \frac{1}{2}\Big(d_{Ch}(\tilde{C}, C) + d_{Ch}(C, \tilde{C})\Big), \quad \text{where }~
d_{Ch}(\tilde{C}, C) = \frac{1}{|C|} \sum_{\vect{X} \in C} \min_{\tilde{\vect{X}} \in \tilde{C}} \| \tilde{\vect{X}} - \vect{X} \|.
\end{equation}
To compute the average per-pixel error between the predicted and ground-truth depth maps $d_{\textrm{depth}}(\hat{D}, D)$, we first normalize the predicted depth to have the same mean and variance as ground truth within the object mask~$\Omega$ in order to deal with the scale ambuiguity of 3D reconstruction under perspective projection.
Then, we compute the mean absolute difference between the the resulting depth maps within~$\Omega$ as
$d_{\textrm{depth}}(\hat{D}, D) = \frac{1}{|\Omega|} \sum_{\vect{y} \in \Omega} | \hat{D}_{\vect{y}} - D_{\vect{y}} |$.

We evaluate on \textbf{Freiburg Cars}~\cite{nima2015unsupervised} dataset, containing videos of cars with ground truth SfM/MVS point clouds and depth maps reporting \dpcl and \ddepth.
In order to prove that \method can learn from independent views of an object category, we construct training batches so that the appearance loss~\eqref{e:crossproj} compares only images of \textit{different} car instances.
We further compare our model to the previously published results on a non-rigid category of human faces, training it on \textbf{CelebA}~\cite{liu2015faceattributes} and testing it on \textbf{Florence 2D/3D Face}~\cite{bagdanov2011florence}.
The latter comes with ground-truth point clouds but no depth maps, so we report~\dpcl for the central portion of the face.
As viewpoints don't vary much in the face data, we also consider \textbf{CUB-200-2011 Birds}~\cite{WahCUB_200_2011}, annotated with 15 semantic 2D keypoints.
It lacks 3D annotations, so we adopt the evaluation protocol of CMR~\cite{kanazawa18learning} and compare against them qualitatively.
We compare to CMR using~\dpcl on 4 categories from \textbf{Pascal3D+}\,\cite{xiang14beyond}, which come with approximate ground-truth shapes obtained by manual CAD model alignment.
We trained and ran HRNet~\cite{Sun2019} to produce input keypoints for evaluation, and also for training where there exists a different dataset to train the detector, i.e.~for cars and faces.
See \Cref{s:details_datasets} for details.

\paragraph{Baseline.}

Our best direct competitor is CMR~\cite{kanazawa18learning}.
For CUB, we use the pre-trained CMR models made available by the authors, and for the other datasets we use their source code to train new models, making sure to use the same train/test splits.
For depth evaluation, we convert the mesh output of CMR into a depth map using the camera parameters estimated by CMR, and for shape evaluation, we convert the mesh into a point cloud by uniformly sampling 30k points on the mesh.

\subsection{Evaluating the canonical map}\label{s:result-corresp}

First, we evaluate the learned canonical map $\Phi_{\vect{y}}(I)$ qualitatively by demonstrating that it captures stable object correspondences.
%
%
In row 2 of \cref{f:tex-transfer}, we overlay image pixels with color-coded 3D canonical embedding vectors $\kappa = \Phi_{\vect{y}}(I)$.
The figure shows that the embeddings are invariant to viewpoint, appearance and deformation.
%
%
Next, we make use of the texture model~\eqref{e:tex-model} to perform texture transfer.
Specifically, given a pair of images $(I_A, I_B)$, we generate an image
$
 I_C(\mathbf{y}) = C(\Phi_\mathbf{y}(I_B);~\vect{\beta}(I_A))
$
that combines the geometry of image~$I_B$ and texture of image~$I_A$.
Row 3 of \Cref{f:tex-transfer} shows texture transfer results for several pairs of images from our benchmark data.

Previous work used keypoint detection error to evaluate the quality of canonical mapping and shape reconstruction.
We argue that it is a biased measure that is easy to satisfy even with degenerate 3D reconstruction or poor canonical mapping outside the keypoints.
We evaluated the percentage of correct keypoints (PCK@0.1) as 85\%, much higher than CSM~\cite{kulkarni19canonical} (48\%) or CMR~\cite{kanazawa18learning} (47\%).
This reflects the fact that \method is non-parametric and is supervised with keypoint locations through the basis~\eqref{e:basis_loss} and reprojection~\eqref{e:repro_loss} losses, so it can easily learn a good keypoint detector.
As we see in row~2 of \cref{f:tex-transfer} though, canonical maps are not discontinuous, thus not overfit to keypoint locations.

\begin{figure}[tb]
\centering
\begin{minipage}{.43\linewidth}
\begin{figure}[H]
\includegraphics[width=\textwidth]{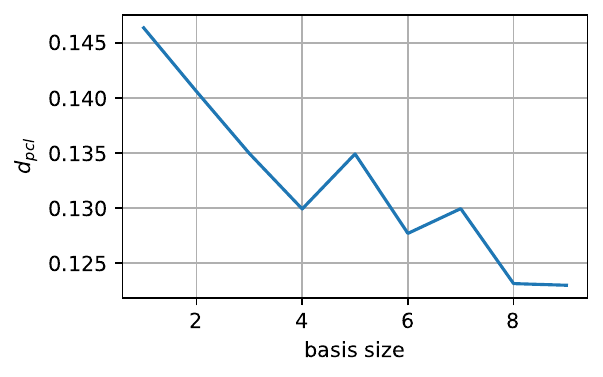}
\vspace{-4.2ex}
\label{f:varbasis}
\caption{\dpcl as a function of basis and shape descriptor size on Freiburg Cars.}
\end{figure}
\end{minipage}\quad
\begin{minipage}{.43\linewidth}
\begin{figure}[H]
\includegraphics[width=\textwidth]{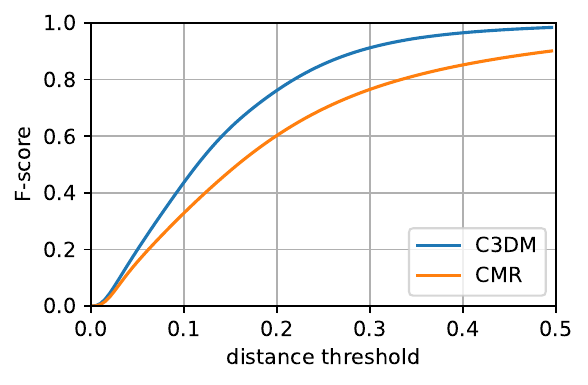}
\vspace{-5ex}
\label{f:fscore}
\caption{F-score of point cloud reconstruction at variable thresholds on Freiburg Cars.}
\end{figure}
\end{minipage}
\end{figure}

\begin{table}[b]
\centering
\begin{minipage}{.57\linewidth}
\input{tab_abl}
\end{minipage}\quad
\begin{minipage}{.4\linewidth}
\input{tab_quant}
\end{minipage}
\end{table}

\subsection{Evaluating 3D reconstructions}\label{s:result-reconst}




\paragraph{Ablation study.}
In~\cref{f:varbasis}, we vary the size of the shape descriptor~$\vect{\alpha}$ and, consequently, the number of blendshapes in~$B$.
The left-most point corresponds to rigid reconstruction.
This sanity check shows that the method can model shape variation rather than predicting generic shape.

In \cref{tab:ablation}, we evaluate the quality of 3D reconstruction by \method trained with different combinations of loss functions.
It shows that each model components improves performance across all metrics and datasets.
The contribution of the appearance loss~\eqref{e:percep_k_loss} is higher for cars, where the keypoints are sparse;
for faces, on the other hand, the network can get far by interpolating between the embeddings of the 98 landmarks even without appearance cues.
The camera-embedding alignment loss~\eqref{e:emb-align} is also more important for cars because of the higher viewpoint diversity.

The last row in \cref{tab:ablation} evaluates the baseline where we replace our representation with a mesh of a fixed topology,
regressing basis vectors at mesh vertices and rasterising their values for image pixels, keeping the same loss functions.
CSM~\cite{kulkarni19canonical} uses a similar procedure to define the cycle consistency loss, with the difference that we rasterise basis vectors~$B(\kappa)$ rather than 3D coordinates~$\vect{X}$.
It allows us to compute the basis matching loss in~\cref{e:basis_loss}, which is defined on keypoints that do not have to correspond to mesh vertices.
On our data, training does not converge to a reasonable shape, probably because the added layer of indirection through mesh makes backpropagation more difficult.

\input{qual-faces-birds}

\paragraph{Comparison with the state-of-the-art.}

\Cref{tab:quant} compares the Chamfer distance~\dpcl and depth error~\ddepth (where applicable) of \method against CMR~\cite{kanazawa18learning}.
On Freiburg Cars and Florence Face, our method attains significantly better results than CMR.
\method produces reasonble reconstructions and generally outperforms CMR on four categories from Pascal3D+ with big lead on chairs.
\Cref{f:fscore} shows that \method attains uniformly higher F-score (as defined by \citet{Tatarchenko2019}) than CMR on Frei.\,Cars.
The visualisations in \cref{f:qual-faces-birds} confirm that \method is better at modelling fine details.

On Freiburg Cars, our method can handle perspective distortions better and is less dependent on instance segmentation failures since it does not have to satisfy the silhouette reprojection loss.
On CelebA, CMR, which relies on this silhouette reprojection loss, produces overly smooth meshes that lack important details like protruding noses.
Conversely, \method leverages the keypoints lifted by C3DPO to accurately reconstruct noses and chins.
On CUB Birds, it is again apparent that \method can reconstruct fine details like beaks.
See \Cref{s:more_results} and videos 
for more visual results.


\section{Conclusions}\label{s:conc}

We have presented \method, a method that learns under weak 2D supervision to densely reconstruct categories of non-rigid objects from single views, establishing dense correspondences between them in the process.
We showed that the model can be trained to reconstruct diverse categories such as cars, birds and human faces, obtaining better results than existing reconstruction methods that work under the same assumptions.
We also demonstrated the quality of dense correspondences by applying them to transfer textures.
The method is still limited by the availability of some 2D supervision (silhouettes and sparse keypoints) at training time.
We aim to remove this dependency in future work.

\clearpage
\section*{Potential broader impact}\label{s:impact}

Our work achieves better image-based 3D reconstruction than the existing technology, which is already available to the wider public.
While we outperform existing methods on benchmarks, however, the capabilities of our algorithm are not sufficiently different to be likely to open new possibilities for misuse.

Our method interprets images and reconstructs objects in 3D.
This is conceivably useful in many applications, from autonomy to virtual and augmented reality.
Likewise, it is possible that this technology, as any other, could be misused.
However, we do not believe that our method is more prone to misuse than most contributions to machine learning.


As for any research output, there is an area of uncertainty on how our contributions could be incorporated in future research work and the consequent impact of that.
We believe that our advances are methodologically significant, and thus we hope to have a positive impact in the community, leading to further developments down the line.
However, it is very difficult to predict the nature of all such possible developments.

\section*{Acknowledgements}\label{s:acknowledgements}
We want to thank Nikhila Ravi for sharing CMR models trained on Pascal3D+ and NeurIPS reviewers for their valuable suggestions.


\newpage\input{appendix}

\end{document}

%% file: fig-teaser.tex
\begin{figure}[t]
\includegraphics[width=\textwidth]{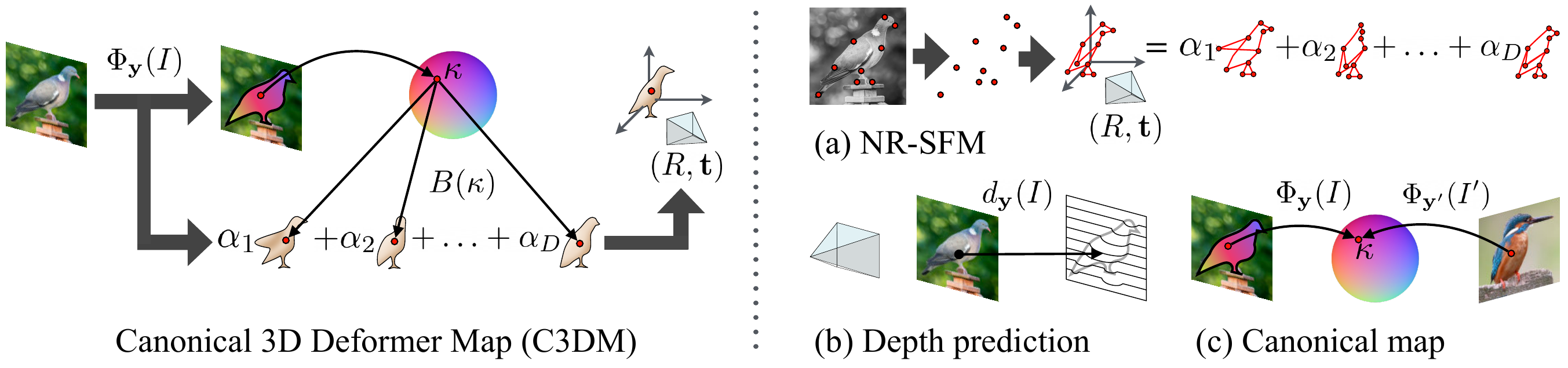}\vspace{-0.25em}
\caption{%
The \method representation (left) associates each pixel~$\mathbf{y}$ of the image~$I$ with a \emph{deformation operator}~$B(\kappa)$, a function of the object canonical coordinates $\kappa = \Phi_\mathbf{y}(I)$.
\method then reconstructs the corresponding 3D point~$\mathbf{X}$ as a function of the global object deformation~$\vect{\alpha}$ and viewpoint~$(R,\mathbf{t})$.
It extends three ideas (right):
(a) non-rigid structure from motion computes a \emph{sparse} parametric reconstruction starting from 2D keypoints rather than an image;
(b) a monocular depth predictor~$d_{\mathbf{y}}(I)$ non-parametrically maps each pixel to its 3D reconstruction but lacks any notion of correspondence;
(c) a canonical mapping~$\Phi_\mathbf{y}(I)$ establishes dense correspondences but does not capture geometry.
}\label{f:teaser}\vspace{-1em}
\end{figure}

%% file: tex-transfer-new.tex
\newcommand{\imw}{.25\textwidth}
\newcommand{\imh}{1.59cm}
\newcommand{\rottextmarg}{0.05cm}

\newcommand{\foldercar}{./images/qual_3d/overview_v18_imnames/freicars-d-sph_nossim_app_v2_nomagnet_00000_MOD.sph_emb_residual=False_MOD.spherical_embedding_for_app=False_M.T.dimout_glob=64}
\newcommand{\folderbird}{./images/qual_3d/overview_v18_imnames/birds-d-repro-cvpr_appmodel_nodilmask_018_smpl_sil_magnet_nrsfms_00009_M.l.loss_vgg=0.0_M.l.ls_shebtocm=1.0_MOD.nrfm_ep_pth=cub_birds_hrnet_orth_b50}

\newcommand{\rottext}[1]{\centering\scriptsize\rotatebox{90}{#1}}

\newcommand{\rotmpage}[1]{
\begin{minipage}[c][\imh][c]{0.1cm}%
\rottext{#1}%
\end{minipage}%
}

\newcommand{\textcolbird}[3]{%
\begin{varwidth}[c]{\imw}\centering%
\noindent\includegraphics[height=\imh]{{\folderbird/#1_im}.png}\\%
\vspace{-0.04cm}%
\noindent\includegraphics[height=\imh]{{\folderbird/#1_uv_image_sph_mix}.png}\\%
\vspace{-0.04cm}%
\noindent\includegraphics[height=\imh]{{\folderbird/#3}.png}\\%
\vspace{-0.04cm}%
\noindent\includegraphics[height=\imh]{{\folderbird/#2_im}.png}\\%
\end{varwidth}}

\newcommand{\textcolcar}[3]{%
\begin{varwidth}[c]{\imw}\centering%
\noindent\includegraphics[height=\imh]{\foldercar/#1_im.png}\\%
\vspace{-0.04cm}%
\noindent\includegraphics[height=\imh]{\foldercar/#1_uv_image_sph_mix.png}\\%
\vspace{-0.04cm}%
\noindent\includegraphics[height=\imh]{\foldercar/#3.png}\\%
\vspace{-0.04cm}%
\noindent\includegraphics[height=\imh]{\foldercar/#2_im.png}\\%
\end{varwidth}} 

\newcommand{\descrrow}[0]{%
\begin{minipage}{0.6cm}%
\centering%
\rotmpage{Target $I_B$}\\%
\rotmpage{Can.~emb. $\Phi(I_B)$}\\%
\rotmpage{Styled $I_C$}\\%
\rotmpage{Texture $I_A$}%
\end{minipage}%
}

\begin{figure*}
\centering%
\descrrow%
\textcolbird{017.Cardinal/Cardinal_0001_17057}{060.Glaucous_winged_Gull/Glaucous_Winged_Gull_0013_44381}{017.Cardinal/Cardinal_0001_17057_im_reenact_060.Glaucous_winged_Gull_Glaucous_Winged_Gull_0013_44381}%
\textcolbird{053.Western_Grebe/Western_Grebe_0042_36035}{180.Wilson_Warbler/Wilson_Warbler_0016_175532}{053.Western_Grebe/Western_Grebe_0042_36035_im_reenact_180.Wilson_Warbler_Wilson_Warbler_0016_175532}%
\textcolbird{130.Tree_Sparrow/Tree_Sparrow_0112_125014}{096.Hooded_Oriole/Hooded_Oriole_0004_91057}{130.Tree_Sparrow/Tree_Sparrow_0112_125014_im_reenact_096.Hooded_Oriole_Hooded_Oriole_0004_91057}%
\textcolcar{037/undistort/images/frame_0000346}{034/undistort/images/frame_0000031}{037/undistort/images/frame_0000346_im_reenact_034_undistort_images_frame_0000031}%
\textcolcar{022/undistort/images/frame_0000178}{036/undistort/images/frame_0000246}{022/undistort/images/frame_0000178_im_reenact_036_undistort_images_frame_0000246}%
\textcolcar{042/undistort/images/frame_0000210}{022/undistort/images/frame_0000019}{042/undistort/images/frame_0000210_im_reenact_022_undistort_images_frame_0000019}\\
\vspace{-1ex}
\caption{\textbf{Canonical mapping and texture transfer for CUB and Freiburg Cars}.
Given a target image $I_B$ (1\textsuperscript{st} row), \method extracts the canonical embeddings $\kappa = \Phi(\vect{y}; I_B)$ (2\textsuperscript{nd} row).
Then, given the appearance descriptor $\vect{\beta}(I_A)$ of a texture image~$I_A$ (4\textsuperscript{th} row), the texture network~$C$ transfers its style to get a styled image
$I_C(\mathbf{y}) = C(\Phi_\mathbf{y}(I_B);~\vect{\beta}(I_A))$
(3\textsuperscript{rd} row), which preserves the geometry of the target image~$I_B$.
Note that we model the texture directly rather than warp the source image, so even the parts occluded in the source image~$I_A$ can be styled (5$^\text{th}$ and 6$^\text{th}$ columns).%
}\label{f:tex-transfer}
\end{figure*}
\let\imw\undefined
\let\imh\undefined
\let\hmrow\undefined
\let\cropbox\undefined
\let\folderbird\undefined
\let\foldercar\undefined

%% file: tab_abl.tex
\begin{tabular}{cccc|c|cc}
\toprule
\multicolumn{4}{c|}{Active Losses $\mathcal{L}$} & 
\multicolumn{1}{c}{Fl.\,Face} & 
\multicolumn{2}{|c}{Frei.\,Cars} \\ \midrule 
$_{\textrm{repro}}$ & 
$_{\textrm{basis}}$ & 
$^{\textrm{min-k}}_{\textrm{percep}}$ &
$^{\textrm{emb-}}_{\textrm{align}}$ & 
\dpcl & \ddepth & \dpcl \\ \midrule
           & \checkmark & \checkmark & \checkmark &   6.582 & 0.548   & 0.247 \\ 
\checkmark &            & \checkmark & \checkmark &   7.406 & 0.550   & 0.462 \\ 
\checkmark & \checkmark &            & \checkmark &   5.647 & 0.361   & 0.141 \\ 
\checkmark & \checkmark & \checkmark &            &   5.592 & 0.498   & 0.186 \\ 
\checkmark & \checkmark & \checkmark & \checkmark &  \textbf{5.574} & \textbf{0.311} & \textbf{0.123} \\ \midrule 
\multicolumn{4}{c|}{interpolation thru mesh}  &   13.721 & 0.596   & 0.182 \\ 
\bottomrule
\end{tabular} 
\caption{\textbf{3D reconstruction accuracy for \method vatiants on cars and faces.} 
We evaluate disabling losses~\eqref{e:repro_loss}, \eqref{e:percep_k_loss}, \eqref{e:emb-align}, and the first term in~\eqref{e:basis_loss}, one by one.}
\label{tab:ablation}




%% file: tab_quant.tex
\begin{tabular}{l|cc}
\toprule
Dataset    & CMR~\cite{kanazawa18learning} & C3DM                      \\ \midrule 
Flo.\,Face  & 13.09      & \textbf{5.57}             \\ \midrule 
Frei.\,Cars & 0.20/0.50  & \textbf{0.12/0.31}        \\ \midrule 
P3D Plane  & 0.022      & \textbf{0.019}            \\
P3D Chair  & 0.049      & \textbf{0.043}            \\
P3D Car    & 0.028      & 0.028                     \\
P3D Bus    & 0.037      & \textbf{0.036}                  \\ \bottomrule
\end{tabular}
\vspace{0.2ex}
\caption{%
\textbf{\dpcl~on Freiburg Cars, Florence Face, and Pascal 3D+}
comparing our method to CMR~\cite{kanazawa18learning}. For Frei.\,Cars, \ddepth~is also reported after slash.%
}\label{tab:quant}

%% file: qual-faces-birds.tex
\newcommand{\imw}{.19\textwidth}
\newcommand{\imwcarcmr}{.14\textwidth}
\newcommand{\cropbox}[1]{\noindent\includegraphics[trim=7cm 7cm 7cm 7cm,clip,width=\imw]{#1}}
\newcommand{\cropboxface}[1]{\noindent\includegraphics[trim=4.5cm 6cm 4.5cm 3cm,clip,width=\imw]{{#1}}}
\newcommand{\cropboxsrcface}[1]{\noindent\includegraphics[trim=0cm 1cm 1.5cm 0.5cm,clip,width=\imw]{#1}}
\newcommand{\cropboxcar}[1]{\noindent\includegraphics[trim=7cm 10cm 7cm 10cm,clip,width=\imw]{#1}}
\newcommand{\cropboxcmr}[1]{\noindent\includegraphics[trim=2cm 2cm 2cm 2cm,clip,width=\imw]{#1}}
\newcommand{\cropboxcmrface}[1]{\noindent\includegraphics[trim=0cm 0cm 0cm 0cm,clip,width=\imw]{#1}}
\newcommand{\cropboxcmrcar}[1]{\noindent\includegraphics[trim=1cm 1cm 1cm 2cm,clip,width=\imw]{#1}}


\newcommand{\folderbird}{./images/qual_3d/overview_v18_imnames/birds-d-repro-cvpr_appmodel_nodilmask_018_smpl_sil_magnet_nrsfms_00009_M.l.loss_vgg=0.0_M.l.ls_shebtocm=1.0_MOD.nrfm_ep_pth=cub_birds_hrnet_orth_b50}
\newcommand{\folderface}{./images/qual_3d/stockfaces}
\newcommand{\foldercar}{./images/qual_3d/overview_v18_imnames/freicars-d-sph_nossim_app_v2_nomagnet_00000_MOD.sph_emb_residual=False_MOD.spherical_embedding_for_app=False_M.T.dimout_glob=64}

\newcommand{\headerentry}[1]{\begin{minipage}{\imw}\centering#1\end{minipage}}

\newcommand{\heading}{
\headerentry{Source\\image}%
\headerentry{Ours,\\view \#1}%
\headerentry{Ours,\\view \#2}%
\headerentry{CMR \cite{kanazawa18learning},\\view \#1}%
\headerentry{CMR \cite{kanazawa18learning},\\view \#2}\\
}

\newcommand{\hmrowbird}[3]{
    \includegraphics[width=\imw]{\folderbird/#1_im.png}%
    \cropbox{\folderbird/#1_im_pcl_full_rdr_#2}%
    \cropbox{\folderbird/#1_im_pcl_full_rdr_#3}%
    \cropboxcmr{\folderbird/#1_vp1.png}%
    \cropboxcmr{\folderbird/#1_vp2.png}%
}

\newcommand{\hmrowface}[3]{
    \cropboxsrcface{\folderface/#1_im.png}%
    \cropboxface{\folderface/#1_pcl_depth_im_0-#2}%
    \cropboxface{\folderface/#1_pcl_depth_im_0-#3}%
    \cropboxcmrface{\folderface/#1_img_pred.png}%
    \cropboxcmrface{\folderface/#1_vp2.png}%
}

\newcommand{\hmrowcar}[3]{
    \includegraphics[width=\imw]{\foldercar/#1_im.png}%
    \cropboxcar{\foldercar/#1_im_pcl_full_rdr_#2}%
    \cropboxcar{\foldercar/#1_im_pcl_full_rdr_#3}%
    \cropboxcmrcar{\foldercar/#1_vp1.png}%
    \cropboxcmrcar{\foldercar/#1_vp3.png}%
}

\begin{figure*}
\centering \scriptsize
\begin{minipage}{0.5\textwidth}\centering
\heading
\hmrowcar{037/undistort/images/frame_0000250}{x0_0_y0_1_z1_0}{x-1_0_y0_1_z-1_0}\\ 
\hmrowcar{042/undistort/images/frame_0000003}{x0_0_y0_1_z1_0}{x-1_0_y0_1_z-1_0}\\
\hmrowface{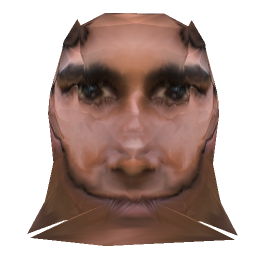}{00}{79}\\
\hmrowface{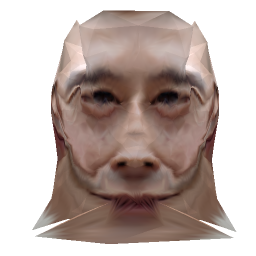}{00}{79}\\
\hmrowface{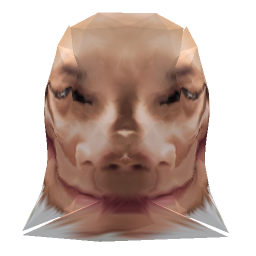}{00}{79}\\
\hmrowface{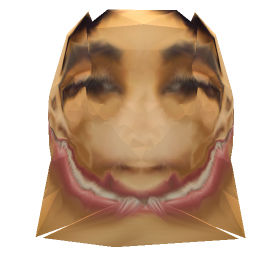}{00}{79}\\
\hmrowface{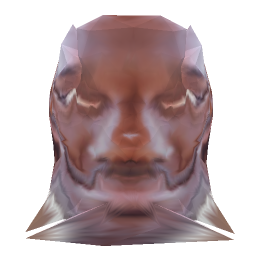}{00}{79}\\
\end{minipage}%
\begin{minipage}{0.5\textwidth}\centering
\heading
\hmrowcar{036/undistort/images/frame_0000295}{x0_0_y0_1_z1_0}{x-1_0_y0_1_z-1_0}\\
\hmrowcar{022/undistort/images/frame_0000022}{x0_0_y0_1_z1_0}{x-1_0_y0_1_z-1_0}\\
\hmrowbird{037.Acadian_Flycatcher/Acadian_Flycatcher_0036_795577}        {x1_0_y0_3_z-1_0}{x0_0_y0_3_z1_0}\\
\hmrowbird{042.Vermilion_Flycatcher/Vermilion_Flycatcher_0030_42523}     {x1_0_y0_3_z-1_0}{x0_0_y0_3_z1_0}\\
\hmrowbird{077.Tropical_Kingbird/Tropical_Kingbird_0036_69939}           {x1_0_y0_3_z-1_0}{x0_0_y0_3_z1_0}\\
\hmrowbird{093.Clark_Nutcracker/Clark_Nutcracker_0047_85630}             {x1_0_y0_3_z-1_0}{x0_0_y0_3_z1_0}\\
\hmrowbird{133.White_throated_Sparrow/White_Throated_Sparrow_0043_128818}{x1_0_y0_3_z-1_0}{x0_0_y0_3_z1_0}\\
\end{minipage}%
\vspace{-1ex}
\caption{\textbf{Visual comparison of the results on Freiburg Cars (top two rows), faces of the authors (left column) and CUB Birds datasets (right column).} 
For each dataset, we show the source image~(1\textsuperscript{st}\,column), \method and CMR reconstructions from the original viewpoint (\emph{view \#1}, 2\textsuperscript{nd}\,and 4\textsuperscript{th}\,columns, respectively) and from an alternative viewpoint (\emph{view \#2}, 3\textsuperscript{rd}\,and 5\textsuperscript{th}\,columns).%
\label{f:qual-faces-birds} }
\end{figure*}

\let\imw\undefined
\let\cropbox\undefined
\let\cropboxcmr\undefined
\let\cropboxcmrface\undefined
\let\cropboxcmrcar\undefined
\let\folderbird\undefined
\let\folderface\undefined
\let\foldercar\undefined
\let\headerentry\undefined
\let\heading\undefined
\let\hmrowbird\undefined
\let\hmrowface\undefined
\let\hmrowcar\undefined

%% file: appendix.tex
\appendix
\renewcommand\thefigure{\Roman{figure}} 


\section{Architecture details} \label{s:arch_details}
\begin{figure}[t]
\centering
\includegraphics[width=0.99\linewidth]{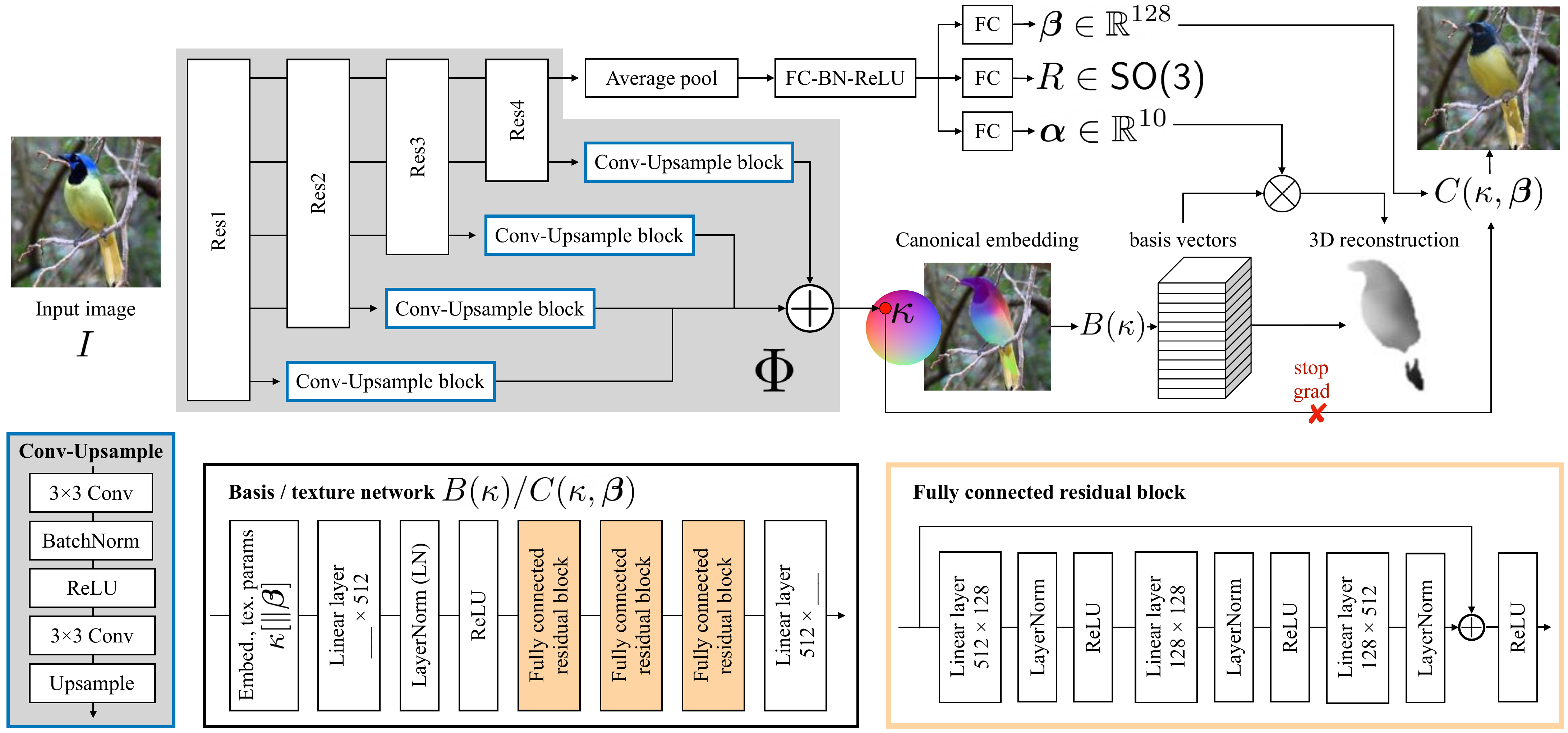}
\caption{\textbf{The detailed architecture of prediction-time \method flow.} All networks share the common ResNet50 backbone.
Camera orientation, shape and texture parameters are regressed from the final residual layer.
The embedding prediction network~$\Phi$ processes outputs of the four residual blocks with the Conv-Upsample subnetwork shown in the left inset, then sums and normalises their outputs to obtain the map of spherical embeddings~$\kappa$.
They are passed through basis and texture networks that share the architecture, which is shown in the middle and right insets.
Finally, the predicted basis vectors are multiplied by shape parameters~$\vect{\alpha}$ to obtain 3D reconstruction of the visible points.%
} \label{f:arch_overview}
\end{figure}

\Cref{f:arch_overview} shows the backbone of our architecture, together with the basis and texture predictors~$B$ and~$C$.
The trunk of \method consists of a Feature Pyramid Network pre-trained on ImageNet.
In more detail, Conv-Upsample blocks are attached to the outputs of each of the Res1, Res2, Res3 and Res4 layers of a ResNet50.
Each Conv-Upsample outputs a tensor with the spatial resolution of the first auxiliary branch that takes Res1 as an input.
The four tensors are then summed and $\ell^2$-normalized in order to produce the canonical embedding tensor~$\kappa$.


The insets of \cref{f:arch_overview} show the architecture of the basis and texture networks $B(\kappa)$ and $C(\kappa, \vect{\beta}(I))$.
The networks follow the C3DPO~\cite{novotny19c3dpo} architecture.
Each of them consists of a fully connected (FC) layer, followed by three fully connected residual blocks (shown in detail in the lower-right inset) and another fully connected layer adapting the output dimensionality.
The LayerNorm layers~\cite{Ba2016} used in these networks only perform $\ell^2$ normalization across channels, without using trainable parameters.
The basis network takes as input the map of 2D canonical embeddings $\kappa$, while the texture network concatenates them with the same texture descriptor~$\vect{\beta}$ to get the 130-dimensional vector for each pixel.
The basis network outputs the 30-dimensional vector for each pixel (10 3-dimensional basis vectors), while the texture network outputs 3D per-pixel colors.

\begin{figure}[tb]
\centering
\includegraphics[width=0.99\linewidth]{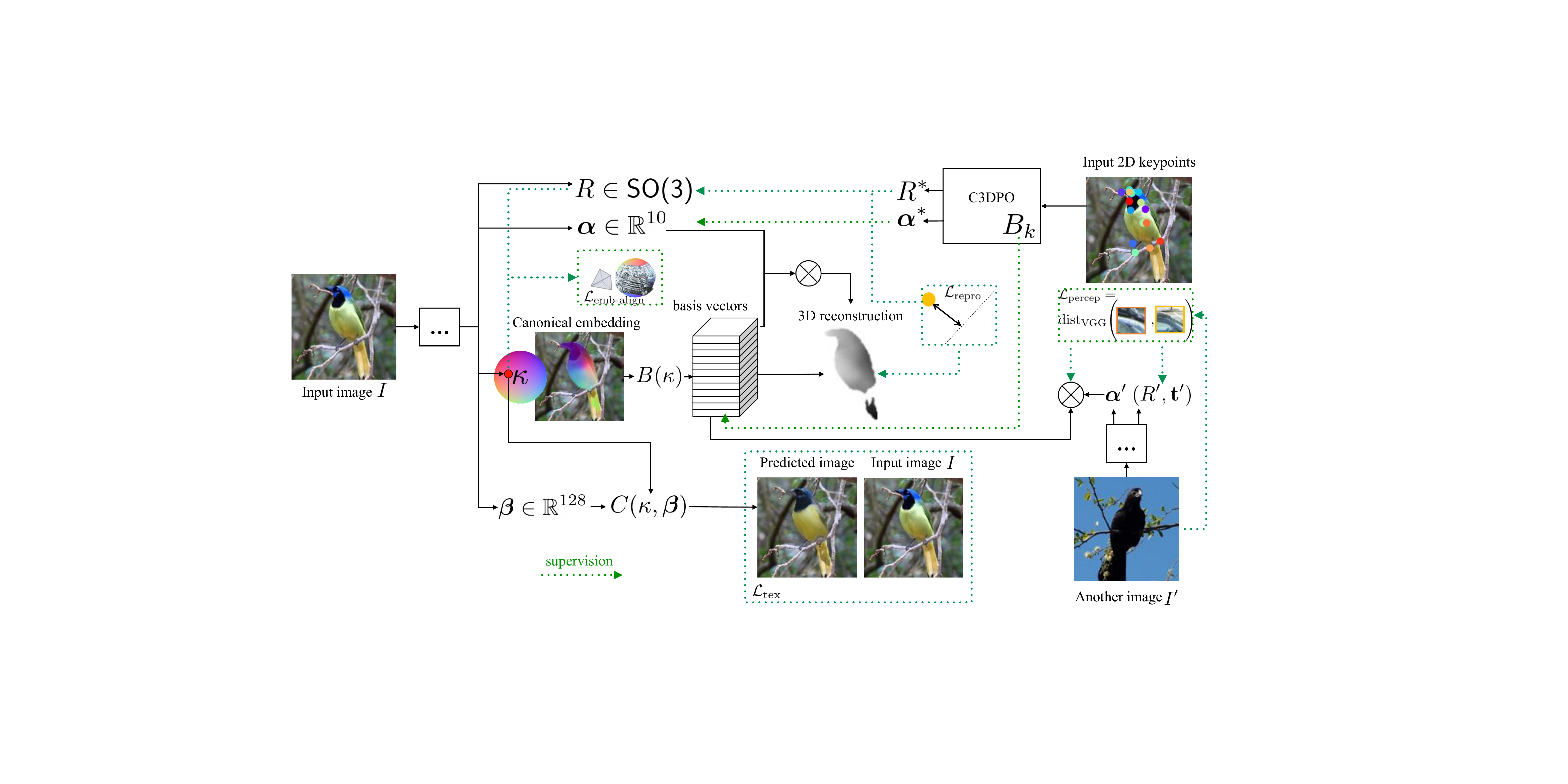}
\caption{\textbf{The training time \method flow,} where the backbones showed in \Cref{f:arch_overview} are collapsed to the boxes with ellipses.
We supervise the predicted basis map with C3DPO bases at keypoint locations.
At training time, we also run C3DPO on 2D keypoints to supervise shape parameters and camera orientation.
Embedding alignment loss acts on the estimated camera orientation and average spherical embeddings.
We project the 3D reconstruction using the estimated camera parameters to define the reprojection consistency loss.
To define the cross-image perceptual consistency loss, we run our network on another image (in practice, the other images in the batch are used) and use its shape and camera parameters to project the estimated basis vectors and compare with that image.
Finally, we supervise the output of the texture model with the original image.%
} \label{f:supp_losses}
\end{figure}

\Cref{f:supp_losses} extends the diagram with the computations specific to the training time.
For supervision, the training also runs C3DPO on 2D keypoints and uses the predictions and bases to define the NR-SFM prior loss~(4) in the maon paper.
The diagram also shows the reprojection consistency loss~(5) in the main paper, cross-image perceptual loss~(7) in the main paper, which requires the viewpoint and shape predictions for other images in the batch, camera-embedding alignment loss~(8) in the main paper, and the texture model loss~\eqref{e:tex-loss}.

\paragraph{Batch sampling.}
In each training epoch, we sample 3000 batches of 10 random images (adding a constraint on Freiburg Cars that they don't come from the same sequence).
We optimize the network using SGD with momentum, starting with learning rate 0.001 and decreasing $10\times$ whenever the objective plateaus.
We stop training after 50 epochs.

Since most datasets are biased in terms of the viewpoints, e.g. birds are less likely to be photographed from the front or back than from the side, we apply inverse propensity correction on the distribution of 1D rotations to ensure uniform coverage.
We correct the distribution of rotations in the horizontal plane only, assuming that the pitch varies less than the azimuth, which is true for most object-centric datasets.
In particular, we first find the upward direction as an eigenvector of the rotation axes extracted from the camera orientations extracted by NR-SFM from the training set:~$\{R^*_n\}$.
Then we compute the azimuth~$a(R^*_n)$ as the rotation component around the estimated upward axis.
The sampling weight for an image~$I_n$ is thus found as~$\big(p(a(R^*_n))\big)^{-1}$, where the distribution~$p$ is approximated by a histogram of 16 bins.
Note that we only need to do this at training time when NR-SFM viewpoint predictions are available; at test time, the networks can take a single image.

To compute the min-k cross-image perceptual loss~(7) in the main paper, we treat the first image~$I$ in the batch as a target and warp the rest of the images using their \textit{estimated} camera and shape parameters~$R(I'), \vect{t}(I'), \vect{\alpha}(I')$.
For each pixel, we average the distances to $k = 6$ closest feature maps as per eq. (7).

\paragraph{Implementation.}
We implemented \method using Pytorch framework. We run training on a single NVidia Tesla V100 GPU with 16 Gb of memory.
Training for full 50 epochs takes around 48 hours.

\paragraph{Runtime analysis}
On a single gpu, the feedforward pass of our network takes one average 0.111 sec per image. 

\section{Details of the photometric and perceptual losses} \label{s:details_percept}
To enforce photometric consistency, we can use the following loss:
\begin{equation}\label{e:photoloss}
\mathcal{L}_\text{photo}(I';\Omega,I)
=
\sum_{\vect{y} \in \Omega}
\| I'(\vect{y}) - I(\vect{y})\|_\epsilon.
\end{equation}
Here $I$ and $I'$~are two images, $\Omega$~is the region of image~$I$ that contains the object (\ie the object mask).

To capture higher-level consistency between images, in particular in the cross-image consistency loss~(6) in the main paper between the target image and warped reference image,
we use \emph{perceptual loss} $\mathcal{L}_\text{percep}$ that compares the activations of a pre-trained neural network~\cite{Zhang2018}.
Specifically, we compute pseudo-Huber loss between the activations of a VGG network, averaged over several layers.
The perceptual loss uses the pretrained VGG-19 network~\cite{Simonyan2015}.
Let~$\Psi_l(I)$ be the layer~$l$ activations of VGG-19 fed by the image~$I$.
We then define the perceptual loss as
\begin{equation}\label{e:perceploss}
\mathcal{L}_\text{percep}(I';\Omega,I)
= \sum_{\vect{y} \in \Omega} \sum_{l \in \{0, 5, 10, 15\}}
\Big\| \textrm{upsample} \big(\Psi_l(I') - \Psi_l(I)\big)[\vect{y}]\Big\|_{\epsilon},
\end{equation}
where $\textrm{upsample}()$ interpolates the feature map to the match the resolution of the network input.

We can now formally define the optimisation problem for the texture model described in Section 3.1.
Given the input image~$I$ and 2D embeddings for all its pixels~$\vect{\kappa}$, it re-produces the image~$I'$ using $I'(\vect{y}) = C(\kappa(\vect{y}); \vect{\beta}(I))$.
The weights of neural networks implementing~$C$ and~$\vect{\beta}$ are found by minimising
\begin{equation}\label{e:tex-loss}
\mathcal{L}_\text{tex}(I';\Omega,I)
=
w_\text{photo} \mathcal{L}^\text{tex}_\text{photo}(I';\Omega,I) + w_\text{percep} \mathcal{L}^\text{tex}_\text{percep}(I';\Omega,I).
\end{equation}
Please note again that the gradients of~$\mathcal{L}_\text{tex}$ are not propagated beyond~$\vect{\kappa}$ to preserve its sole dependence on geometry.

\section{Camera models and ray-projection loss} \label{s:details_camera}

\paragraph{Camera models.}
We have to define a camera model $\pi : \mathbb{R}^3 \rightarrow \mathbb{R}^2$ mapping 3D points in the coordinate frame of the camera to 2D image points in order to compute reprojection and photometric losses.
If the camera calibration is unknown (as in CelebA, Florence Face, CUB, Pascal 3D+ datasets), we use an \emph{orthographic camera}
$
\pi(\vect{X}) = [x_1,x_2]^\top
$
where $\vect{X}=[x_1,x_2,x_3]^\top$.
In this case, we also set $\vect{t}=0$ as translation can be removed by centering the 2D data~\cite{novotny19c3dpo} in pre-processing.

If the camera calibration is known (in Freiburg Cars), we can also use a more accurate \emph{perspective camera model} instead:
\begin{equation}\label{e:perspective}
\pi(\vect{X}) = \frac{f}{x_3}
\begin{bmatrix}
x_1\\
x_2\\
\end{bmatrix},
\end{equation}
where~$f$ is the focal length.

Further to Section 3.1, here, we describe additional implementation details that were important for the success of the perspective projection model on the Freiburg Cars dataset.

\paragraph{Ray-projection loss}
For perspective model, we have also found an improvement that significantly stabilizes the C3DPO algorithm that we use to constrain \method.
The idea is to modify reprojection loss to measure, instead of the distance between 2D projections $\vect{y}$ and $\hat{\vect{y}}$,
the distance of the 3D point $\vect{X}(\vect{y})$ to the line passing through $\vect{y}$ and the camera center.
The advantage is removing the division embedded in the perspective projection equation~\eqref{e:perspective}.

In order to minimize the reprojection error~(5) in the main paper under the perspective projection model,
a na\"{i}ve implementation would minimize the following perspective re-projection loss:
\begin{equation}\label{e:repro_loss_persp}
\mathcal{L}_\text{repro}^{\text{persp}}(\Phi; \Omega, I) =
\sum_{\vect{y}\in\Omega} \big\| \pi_{I,\vect{0}}(\vect{X}_{R,\vect{t}}(\vect{y}))
-
\vect{y}
\big\|_\epsilon,
\end{equation}
where $\vect{X}_{R,\vect{t}}(\vect{y}) = R \vect{X}(\vect{y}) + \vect{t}$ is the 3D point extracted from pixel $\vect{y}$ and expressed in the coordinate frame of the camera of the image $I_n$.
Unfortunately, we found that the division in the perspective projection formula $\pi_{I,\vect{0}} = \frac{f}{x_3} [ x_1 ~ x_2 ]^\top$ leads to unstable training.
This is due to exploding gradient magnitudes caused by 3D points~$\vect{X}$ predicted to lie too close to the camera projection plane.
While this could be extenuated by clamping the points to lie in a safe distance from the camera plane, due to the non-linearity of the projection gradient, the re-projection loss~\eqref{e:repro_loss_persp} still would not converge stably.

In order to remove the gradient non-linearity, we alter the re-projection loss to the \emph{ray-projection} loss:
\begin{equation}\label{e:repro_loss_ray}
\mathcal{L}_\text{repro}^{\text{ray}}(\Phi; \Omega, I) =
\sum_{\vect{y}\in\Omega}
\Big\|
\vect{X}_{R,\vect{t}}(\vect{y})
-
\big[\vect{r}(\vect{y})^\top \vect{X}_{R,\vect{t}}(\vect{y})\big] \vect{r}(\vect{y})
\Big\|_\epsilon,
\end{equation}
where $\vect{r}(\vect{y})$
stands for the direction vector of the projection ray passing through the pixel $\vect{y}$ in the image $I$: 
$$
\vect{r}(\vect{y}) =
\frac{K^{-1} [y_1~y_2~1]^\top}
{\| K^{-1} [y_1~y_2~1]^\top\|},
$$
where $K$ is the instrinsic camera calibration matrix. Intuitively, \cref{e:repro_loss_ray} minimizes the orthogonal distance between the the estimated point~$\vect{X}_{R,\vect{t}}(\vect{y})$ and its projection on the ground truth projection ray~$\vect{r}(\vect{y})$.
We notice that \cref{e:repro_loss_ray} is linear in $\vect{X}_{R,\vect{t}}$ on infinity and quadratic in the compact region around the optimum, hence the magnitude of the gradient is bounded from above.
We found this addition important for convergence of \method.

\paragraph{Perspective projection for C3DPO}
In order to optimize \cref{e:repro_loss_ray}, a C3DPO model \cite{novotny19c3dpo} trained using the perspective projection model is required.
Since the original C3DPO codebase only admits orthographic cameras, we will describe additions to the pipeline that enable training a perspective model on Freiburg Cars.

C3DPO optimizes a combination of canonicalization and reprojection losses.
To this end, we replace the original C3DPO reprojection loss (eq. (4) in \cite{novotny19c3dpo}) with the ray-projection loss~\eqref{e:repro_loss_ray}.
Additionally, unlike in the orthographic case, one has to determine the full 3DoF position of the camera w.r.t.\,the object coordinate frame.
While it is possible to let C3DPO predict translation as an additional output of the network, we avoid over-parametrization of the problem by estimating camera translation as a solution to a simple least-squares problem.

In more detail, we exploit the locally quadratic form of the ray-projection loss and formulate the translation estimation problem that allows for a closed-form solution.
Assuming that C3DPO, given a list of input 2D landmarks $\vect{y}_1, ..., \vect{y}_K$, predicts a camera rotation matrix $R$, the translation can be obtained as a solution to the following problem:
$$
\vect{t^*} = \text{arg} \text{min}_{\vect{t}}
\sum_{i=1}^K
\big\|
\vect{X}_{R,\vect{t}}(\vect{y}_k)
-
    \vect{r}(\vect{y}_k)^\top \vect{X}_{R,\vect{t}}(\vect{y}_k)
 \vect{r}(\vect{y}_k)
\big\|^2.
$$
After a few mathematical manipulations, we arrive at the following closed-form expression for~$\vect{t^*}$:
\begin{equation}\label{e:persp_trans_sol}
\vect{t^*} =
\bigg[\sum_{k=1}^K (I - \Gamma_k)\bigg]^{-1}
\bigg[\sum_{k=1}^K (\Gamma_k - I) \vect{X}_{R,\vect{0}} \bigg],
\end{equation}
where $\Gamma_k = \vect{r}(\vect{y}_k) \vect{r}(\vect{y}_k)^\top$ is an outer product of $\vect{r}(\vect{y}_k)$ with itself.
Using \cref{e:persp_trans_sol}, we can estimate the camera translation online during the SGD iterations of the C3DPO optimization.
Note that the matrix inverse in \cref{e:persp_trans_sol} is not an issue because of the small size of the matrix being inverted~(3$\times$3)
and the possibility to backpropagate through matrix inversion using modern automatic differentiation frameworks (PyTorch).

\section{Rotation loss} \label{s:details_rotation}
We use the distance between rotation matrices~$d_{\epsilon}(R, R^*)$ as part of the loss~(4) in the main paper.
We aim to penalise large angular distance, while avoiding the exploding gradients of inverse trigonometric functions.
First, we note that the relative rotation can be computed as~$R^\top R^*$.
Next, converting it to the axis-angle representation lets us compute the angular component as $\theta = \arccos \big(\frac{1}{2}(\textrm{Tr}(R^{\top}R^*)-1)\big)$.
Using the fact that $\arccos$ is monotonically decreasing, we strip it and apply an affine transform to make sure the loss achieves the minimum at 0:
\begin{equation}\label{e:depsilon}
d_{\epsilon}(R, R^*) = 1 - \cos \theta = \frac{3 - \textrm{Tr}(R^{\top}R^*)}{2}.
\end{equation}

\section{Datasets} \label{s:details_datasets}

\paragraph{Freiburg Cars \textbf(FrC).}
In order to test our algorithm in a low-noise setting, we consider the Freiburg cars dataset~\cite{nima2015unsupervised}\footnote{\url{https://github.com/lmb-freiburg/unsup-car-dataset}} containing walkaround videos of 52 cars.
While this dataset contains videos of the cars, in order to test the ability of the photometric loss~(7) in the main paper to reconstruct objects even if the views are independent,
we pair each pivot image~$I$ with a selection of other images~$\mathcal{P}_I$ extracted from \emph{different} video sequences.

Following \citet{novotny18capturing,novotny17learning}, we set out 5 sequences for validation (indexed 22, 34, 36, 37, 42).
The training set contains 11,162 training frames and 1,427 validation frames.
For evaluation, we also use their ground-truth 3D point clouds, but we only retain the 3D points that, after being projected into each image of a given test sequence, fall within the corresponding segmentation mask.
Each point cloud is further normalized to zero-mean and unit variance along the 3 coordinate axes.
Please refer to~\cite{novotny18capturing} for details.

As an input to our method, we use the pre-trained Mask R-CNN of~\cite{he2017mask} to extract the segmentation masks and the HRNet~\cite{sun2019deep} trained on PASCAL 3D+~\cite{xiang2014beyond} to extract the 2D keypoints.
Hence, all inputs to our method are extracted automatically.
We excluded the frames where a car was detected with a confidence below a threshold.

We report the Chamfer distance~\dpcl~between the ground truth and the predicted point clouds after rigid alignment via ICP~\cite{besl1992method}.
The point cloud predictions are obtained as explained in the \emph{Benchmarks} section of the main text, with $|\mathcal{B}|=30\mathrm{k}$.
Furthermore, we evaluate the quality of our depth predictions
by measuring the average depth distance \ddepth~between the point cloud formed by un-projecting the predicted depth map and the visible part of the ground truth point cloud.

\paragraph{CelebA and Florence faces (\textbf{FF}).}

The FrC dataset contains deformation between object instances, but each object itself is rigid.
In order to compare the ability of our method to handle instance-level non-rigid deformations with the CMR's, we also run the method on images of human faces; in particular, we train our algorithm on the training set of CelebA dataset~\cite{liu2015faceattributes}\footnote{\url{http://mmlab.ie.cuhk.edu.hk/projects/CelebA.html}} containing 161,934 face images and test it on the Florence 2D/3D Face dataset~\cite{bagdanov2011florence}\footnote{\url{http://www.micc.unifi.it/masi/research/ffd/} \copyright Copyright 2011--2019 MICC --- Media Integration and Communication Center, University of Florence. The Florence 2D/3D Face Dataset.}.
The latter contains videos of 53 people and their ground truth 3D meshes, which we can use to assess the quality of our 3D reconstructions.
Following a standard practice, we crop each 3D mesh to retain points that lie within 100mm distance from the nose tip.
We extract 98 semantic keypoints for each training and test face using the pre-trained HRNet detector of~\cite{Sun2019}.

For evaluation on FF, five frames are uniformly sampled from each test sequence.
We then use our network to reconstruct each test face in 3D and evaluate \dpcl~after ICP alignment.
Since the extent of the predicted face differs from the ground truth, we first pre-align the prediction by registering a 3D crop that covers the convex hull of the 98 semantic keypoints.
The 100mm nose-tip crop is then extracted from the pre-aligned mesh and is aligned for the second time.
\ddepth~is not reported for FF since the dataset does not contain ground truth per-frame depth.

\paragraph{CUB-200-2011 Birds.}

We evaluate our method qualitatively on the CUB Birds dataset~\cite{WahCUB_200_2011}\footnote{\url{http://www.vision.caltech.edu/visipedia/CUB-200-2011.html}}, which consists of 11,788 still images of birds belonging to 200 species.
Each image is annotated with 15 semantic keypoints. As done in~\cite{novotny19c3dpo}, for evaluation we use detections of a pre-trained HRNet.
The dataset is challenging mainly due to significant shape variations across bird species, in addition to instance-level articulation.
Since there is no 3D ground truth for that dataset, we qualitatively compare the quality of 3D reconstruction to the ones of CMR~\cite{kanazawa18learning}.
We also use the same training/validation split as CMR.

\paragraph{Pascal3D+.}
We provide additional comparison to CMR on four categories of Pascal3D+\,\cite{xiang14beyond}\footnote{\url{https://cvgl.stanford.edu/projects/pascal3d.html}}:
aeroplane, consisting of 1194 training and 1135 test images, bus (674 training / 657 test), car (2765 training / 2713 test), and chair (650 training / 666 test).
It has been manually annotated by rigidly aligning one of category-specific CAD models, so the annotation has noisy and biased shape and pose.
Since the original CMR codebase contains models for only two classes, we trained CMR models on all considered classes ourselves (using their codebase) and test on the corresponding validation sets.
We report only \dpcl, since the depth maps obtained by projecting with noisy cameras are unreliable.

\section{Hyperparameters used in experiments} \label{s:details_hyperp}
To sum up, during training, we optimize the following weighted sum of loss functions:
\begin{equation}\label{e:all_losses}
\begin{split}
\mathcal{L}(\Phi,B,\vect{\alpha},R,\vect{t}, \hat{I};\Omega,I,\mathcal{P}_I,\mathcal{A}^*) =~ &
w_\text{pr} \mathcal{L}_\text{pr}(\Phi,B,\vect{\alpha},R;I,Y,\mathcal{A}^*) +\\
& w_\text{repro} \mathcal{L}_\text{repro}(\Phi, B, \vect{\alpha}, R; \Omega, I) +\\
& w^\text{min-k}_\text{percep} \mathcal{L}^\text{min-k}_\text{percep}(\Phi,B,\vect{\alpha},R,\vect{t};\Omega,I,\mathcal{P}_I) +\\
& w_\text{emb-align} \mathcal{L}_\text{emb-align}(\Phi,R;\Omega,I) +\\
& w_\text{mask} \mathcal{L}_\text{mask}(B, \vect{\alpha}, R, \vect{t}; \Omega) +\\
& \mathcal{L}_\text{tex}(\hat{I};\Omega,I).
\end{split}
\end{equation}

We set most weights such that the corresponding term has a magnitude of about 1 in the beginning of training.
We set $w_\text{pr} = 1$, $w_{\vect{\alpha}} = 1$, $w_\text{repro} = 1$ for the perspective camera model and $w_\text{repro} = 0.01$ for the orthographic one, where the error is measured in pixels rather than world units.
For the components of texture loss, we set $w^\text{tex}_\text{photo} = 1$, and $w^\text{tex}_\text{percep} = 0.1$.
Likewise, we set the weight for the geometry perceptual loss $w^\text{min-k}_\text{percep} = 0.1$.
We ran grid search for the camera-related parameters within the following ranges:
$w_{R} \in \{1, 10\}$, and $w_\text{emb-align} \in \{1, 10\}$.
We enable~$\mathcal{L}_\text{mask}$ for CUB Birds, Faces, and Pascal3D+ aeroplanes and chairs with weight $w_\text{mask} = 1$.

\section{Additional qualitative results} \label{s:more_results}

\Cref{f:qual_cars_appendix,f:qual_birds_appendix} contain additional single-view reconstruction results.
We can see that \method is robust to occlusions and instance segmentation failures: the 3D shape is reasonably completed in those cases.
Furthermore, \cref{f:tex_transfer_appendix_bird,f:tex_transfer_appendix_car} have been populated with supplemental texture transfer results.
Note that all images are taken from the test set, and images from the same FrC sequence do not co-occur in training and test sets.
We also invite the readers to watch the attached videos of the rendered reconstructions to better evaluate 3D reconstruction quality.



\input{qual_suppl}

\input{tex-transfer-suppl}

\input{tex-transfer-suppl-car}

\cleardoublepage
\let\oldthebibliography\thebibliography
\let\endoldthebibliography\endthebibliography
\renewenvironment{thebibliography}[1]{
  \begin{oldthebibliography}{#1}
    \setlength{\itemsep}{0em}
    \setlength{\parskip}{0em}
}
{
  \end{oldthebibliography}
}
{\small\bibliographystyle{plainnat}\bibliography{refs}}


%% file: qual_suppl.tex
\newcommand{\imh}{2.5cm}

\graphicspath{{./images/result-bird/}}

\newcommand{\hmrow}[3]{%
    \includegraphics[height=\imh]{#1_im.png}%
    \includegraphics[height=\imh]{#1_uv_image_sph_mix.png}%
    \includegraphics[trim=3.3cm 3.3cm 3.3cm 3.3cm,clip,height=\imh]{#1_im_pcl_full_rdr_#2.png}%
    \includegraphics[trim=3.3cm 3.3cm 3.3cm 3.3cm,clip,height=\imh]{#1_im_pcl_full_rdr_#3.png}%
}

\newcommand{\figcapt}[1]{
\textbf{#1}. Columns: input image; canonical mapping; 3D reconstruction with the reconstructed texture from two viewpoints.
}

\begin{figure*}[h!]
\centering
\hmrow{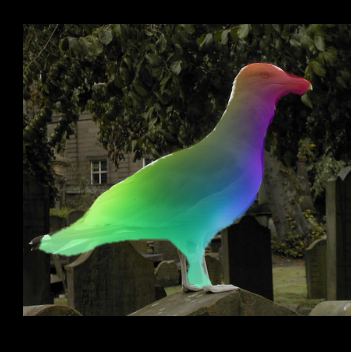}{x1_0_y0_3_z-1_0}{x1_0_y0_3_z1_0}
\hmrow{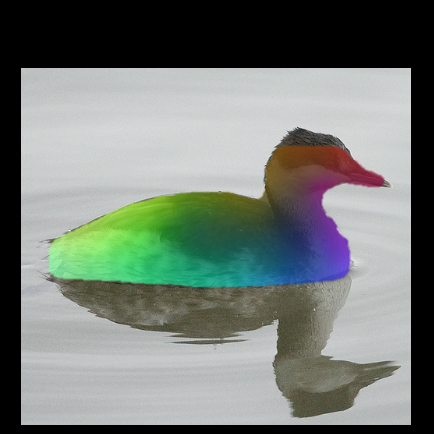}{x-1_0_y0_3_z-1_0}{x-1_0_y0_3_z1_0}
\hmrow{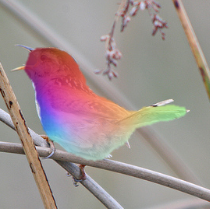}{x0_0_y0_3_z-1_0}{x1_0_y0_3_z-1_0}
\hmrow{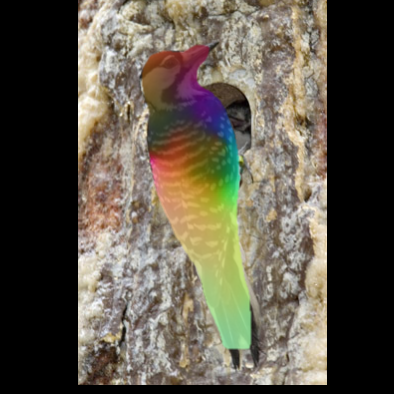}{x-1_0_y0_3_z-1_0}{x1_0_y0_3_z-1_0}
\hmrow{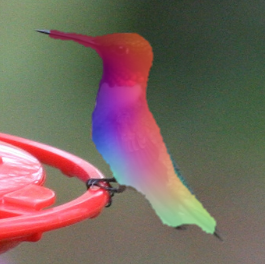}{x-1_0_y0_3_z-1_0}{x-1_0_y0_3_z1_0}
\hmrow{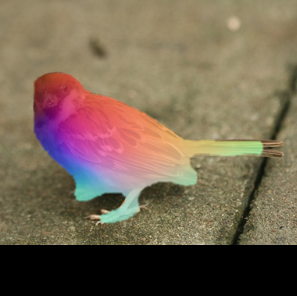}{x0_0_y0_3_z1_0}{x1_0_y0_3_z1_0}
\hmrow{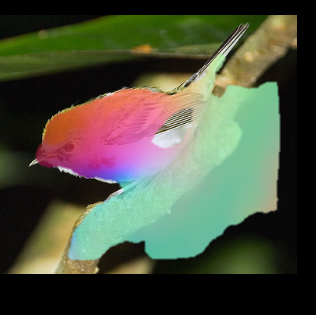}{x0_0_y0_3_z-1_0}{x1_0_y0_3_z-1_0}
\caption{\figcapt{Additional single-view reconstruction results on images from the test sequences of CUB Birds} } \label{f:qual_birds_appendix}
\end{figure*}

\graphicspath{{./images/result-car/}}

\begin{figure*}[h!]
\centering
\hmrow{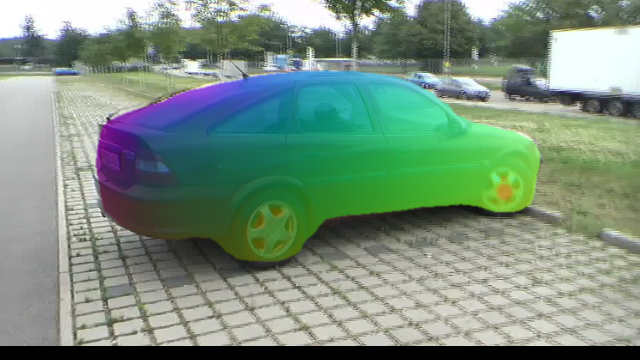}{x-1_0_y0_3_z1_0}{x1_0_y0_3_z1_0}
\hmrow{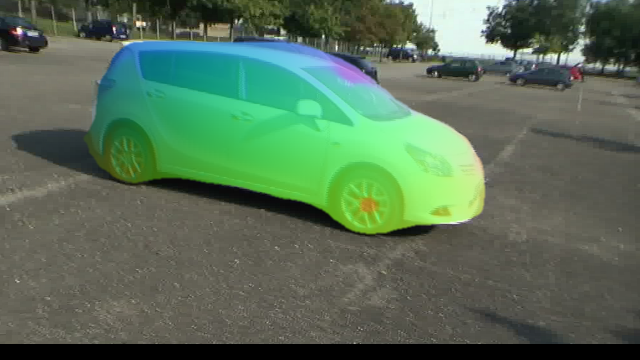}{x0_0_y0_3_z-1_0}{x1_0_y0_3_z1_0}
\hmrow{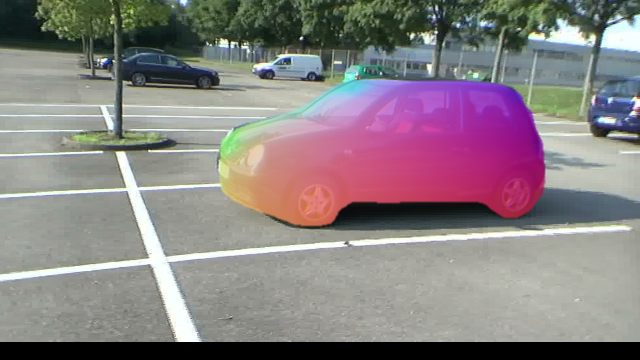}{x-1_0_y0_3_z1_0}{x1_0_y-0_3_z1_0}
\hmrow{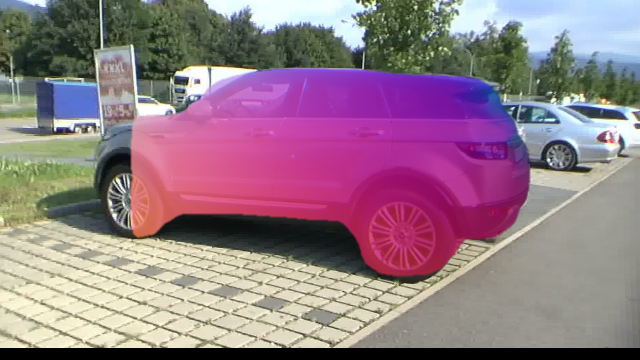}{x-1_0_y0_1_z1_0}{x0_0_y0_3_z1_0}
\hmrow{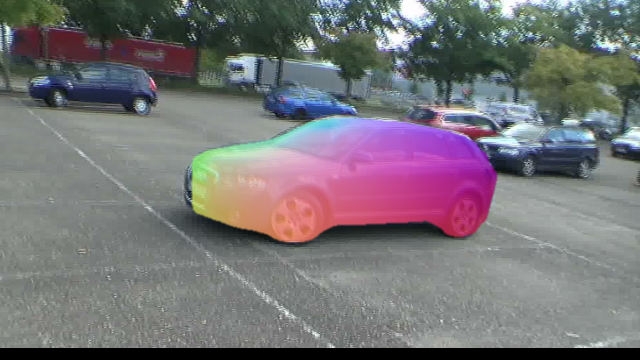}{x-1_0_y0_3_z1_0}{x0_0_y0_3_z1_0}
\caption{\figcapt{Additional single-view reconstruction results on images from the test sequences of Freiburg Cars} } \label{f:qual_cars_appendix}
\end{figure*}


\let\imh\undefined
\let\figcapt\undefined
\let\hmrow\undefined

%% file: tex-transfer-suppl.tex
\newcommand{\imh}{3cm}
\newcommand{\imw}{3cm}

\makeatletter
\def\maxheight#1{\ifdim\Gin@nat@height>#1 #1\else\Gin@nat@height\fi}
\makeatother

\graphicspath{{./images/tex-transfer-supp/}}

\newcommand{\figcapt}[1]{
\textbf{Canonical mapping and texture transfer for #1.}
Given a target image $I_B$ (1\textsuperscript{st} row), \method extracts the canonical embeddings $\kappa = \Phi(\vect{y}; I_B)$ (2\textsuperscript{nd} row).
Then, given the appearance descriptor $\vect{\beta}(I_A)$ of a texture image~$I_A$ (4\textsuperscript{th} row), the texture network~$C$ transfers its style to get a styled image
$I_C(\mathbf{y}) = C(\Phi_\mathbf{y}(I_B);~\vect{\beta}(I_A))$
(3\textsuperscript{rd} row), which preserves the geometry of the target image~$I_B$.%
}

\newcommand{\fixhrow}[3]{%
\noindent\includegraphics[height=\imh]{#3/#1_im.png}&
\noindent\includegraphics[height=\imh]{#3/#1_uv_image_sph_mix.png}&
\noindent\includegraphics[height=\imh]{#3/#1_im_reenact_#2.png}&
\noindent\includegraphics[height=\imh]{#3/#2_im.png}\\
}



\begin{figure*}
\def\arraystretch{0.}
\centering \footnotesize
\begin{tabular}{@{}c@{}c@{}c@{}l@{}}
\vspace{0.2cm} Target image $I_B$&%
Can.~emb. $\Phi(I_B)$&%
Styled image $I_C$&%
Texture image $I_A$\\
\fixhrow{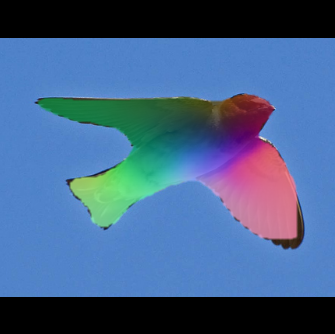}{Tropical_Kingbird_0036_69939}{bird}\\
\fixhrow{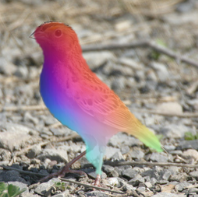}{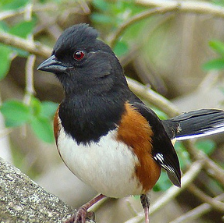}{bird}\\
\fixhrow{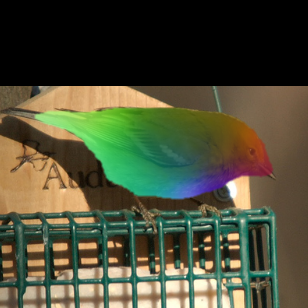}{Acadian_Flycatcher_0036_795577}{bird}\\
\fixhrow{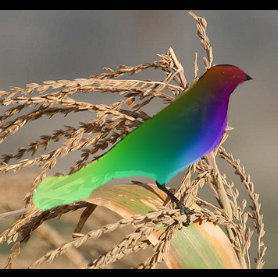}{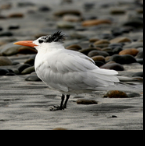}{bird}\\
\fixhrow{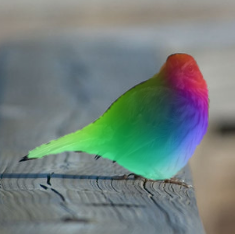}{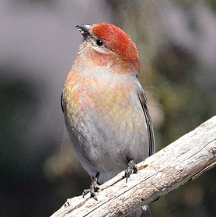}{bird}\\
\end{tabular}
\caption{\figcapt{CUB}} \label{f:tex_transfer_appendix_bird}
\end{figure*}
    
\let\imh\undefined
\let\imw\undefined
\let\figcapt\undefined
\let\fixhrow\undefined

%% file: tex-transfer-suppl-car.tex
\newcommand{\imh}{2cm}
\newcommand{\imw}{3cm}

\makeatletter
\def\maxheight#1{\ifdim\Gin@nat@height>#1 #1\else\Gin@nat@height\fi}
\makeatother

\newcommand{\foldercar}{./images/qual_3d_supp/overview_v18_imnames/freicars-d-sph_nossim_app_v2_nomagnet_00000_MOD.sph_emb_residual=False_MOD.spherical_embedding_for_app=False_M.T.dimout_glob=64/}

\newcommand{\figcapt}[1]{
\textbf{Canonical mapping and texture transfer for #1.}
Given a target image $I_B$ (1\textsuperscript{st} row), \method extracts the canonical embeddings $\kappa = \Phi(\vect{y}; I_B)$ (2\textsuperscript{nd} row).
Then, given the appearance descriptor $\vect{\beta}(I_A)$ of a texture image~$I_A$ (4\textsuperscript{th} row), the texture network~$C$ transfers its style to get a styled image
$I_C(\mathbf{y}) = C(\Phi_\mathbf{y}(I_B);~\vect{\beta}(I_A))$
(3\textsuperscript{rd} row), which preserves the geometry of the target image~$I_B$.%
}

\newcommand{\fixhrow}[3]{%
\noindent\includegraphics[height=\imh]{\foldercar/#1_im.png}&%
\vspace{-0.04cm}%
\noindent\includegraphics[height=\imh]{\foldercar/#1_uv_image_sph_mix.png}&%
\vspace{-0.04cm}%
\noindent\includegraphics[height=\imh]{\foldercar/#3.png}&%
\vspace{-0.04cm}%
\noindent\includegraphics[height=\imh]{\foldercar/#2_im.png}
\\%
}

\begin{figure*}
\def\arraystretch{0.}
\centering \footnotesize
\begin{tabular}{@{}c@{}c@{}c@{}l@{}}
\vspace{0.2cm} Target image $I_B$&%
Can.~emb. $\Phi(I_B)$&%
Styled image $I_C$&%
Texture image $I_A$\\
\fixhrow{022/undistort/images/frame_0000042}{037/undistort/images/frame_0000313}{022/undistort/images/frame_0000042_im_reenact_037_undistort_images_frame_0000313}
\fixhrow{022/undistort/images/frame_0000226}{034/undistort/images/frame_0000181}{022/undistort/images/frame_0000226_im_reenact_034_undistort_images_frame_0000181}
\fixhrow{022/undistort/images/frame_0000280}{034/undistort/images/frame_0000243}{022/undistort/images/frame_0000280_im_reenact_034_undistort_images_frame_0000243}
\fixhrow{034/undistort/images/frame_0000114}{036/undistort/images/frame_0000297}{034/undistort/images/frame_0000114_im_reenact_036_undistort_images_frame_0000297}
\fixhrow{036/undistort/images/frame_0000070}{042/undistort/images/frame_0000151}{036/undistort/images/frame_0000070_im_reenact_042_undistort_images_frame_0000151}
\fixhrow{037/undistort/images/frame_0000018}{022/undistort/images/frame_0000022}{037/undistort/images/frame_0000018_im_reenact_022_undistort_images_frame_0000022}
\fixhrow{037/undistort/images/frame_0000232}{036/undistort/images/frame_0000043}{037/undistort/images/frame_0000232_im_reenact_036_undistort_images_frame_0000043}
\fixhrow{037/undistort/images/frame_0000250}{022/undistort/images/frame_0000180}{037/undistort/images/frame_0000250_im_reenact_022_undistort_images_frame_0000180}
\fixhrow{042/undistort/images/frame_0000174}{036/undistort/images/frame_0000285}{042/undistort/images/frame_0000174_im_reenact_036_undistort_images_frame_0000285}
\fixhrow{042/undistort/images/frame_0000255}{022/undistort/images/frame_0000023}{042/undistort/images/frame_0000255_im_reenact_022_undistort_images_frame_0000023}
\end{tabular}
\caption{\figcapt{Freiburg cars}} \label{f:tex_transfer_appendix_car} 
\end{figure*}

\let\imw\undefined
\let\imh\undefined
\let\fixhrow\undefined
\let\foldercar\undefined

%% file: main.bbl
\begin{thebibliography}{81}
\providecommand{\natexlab}[1]{#1}
\providecommand{\url}[1]{\texttt{#1}}
\expandafter\ifx\csname urlstyle\endcsname\relax
  \providecommand{\doi}[1]{doi: #1}\else
  \providecommand{\doi}{doi: \begingroup \urlstyle{rm}\Url}\fi

\bibitem[Agudo and Moreno-Noguer(2017)]{agudo2017dust}
Antonio Agudo and Francesc Moreno-Noguer.
\newblock Dust: Dual union of spatio-temporal subspaces for monocular multiple
  object 3d reconstruction.
\newblock In \emph{Proc. {CVPR}}, 2017.

\bibitem[Agudo and Moreno-Noguer(2018)]{agudo2018deformable}
Antonio Agudo and Francesc Moreno-Noguer.
\newblock Deformable motion {3D} reconstruction by union of regularized
  subspaces.
\newblock In \emph{Proc. {ICIP}}, 2018.

\bibitem[Agudo et~al.(2018)Agudo, Pijoan, and Moreno-Noguer]{agudo2018image}
Antonio Agudo, Melcior Pijoan, and Francesc Moreno-Noguer.
\newblock Image collection pop-up: {3D} reconstruction and clustering of rigid
  and non-rigid categories.
\newblock In \emph{Proc. {CVPR}}, 2018.

\bibitem[Akhter et~al.(2009)Akhter, Sheikh, Khan, and
  Kanade]{akhter2009nonrigid}
Ijaz Akhter, Yaser Sheikh, Sohaib Khan, and Takeo Kanade.
\newblock Nonrigid structure from motion in trajectory space.
\newblock In \emph{Proc. {NIPS}}, 2009.

\bibitem[Akhter et~al.(2011)Akhter, Sheikh, Khan, and
  Kanade]{akhter2011trajectory}
Ijaz Akhter, Yaser Sheikh, Sohaib Khan, and Takeo Kanade.
\newblock Trajectory space: A dual representation for nonrigid structure from
  motion.
\newblock \emph{{PAMI}}, 33\penalty0 (7):\penalty0 1442--1456, 2011.

\bibitem[Anguelov et~al.(2005)Anguelov, Srinivasan, Koller, Thrun, Rodgers, and
  Davis]{anguelov05scape}
D.~Anguelov, P.~Srinivasan, D.~Koller, S.~Thrun, J.~Rodgers, and J.~Davis.
\newblock {SCAPE}: shape completion and animation of people.
\newblock In \emph{{ACM} Trans. on Graphics}, 2005.

\bibitem[Ayache and Faverjon(1986)]{ayache86efficient}
N.~Ayache and B.~Faverjon.
\newblock Efficient registration of stereo images by matching graph description
  of edge segments.
\newblock Technical Report 559, {INRIA}, 1986.

\bibitem[Ba et~al.(2016)Ba, Kiros, and Hinton]{Ba2016}
Jimmy~Lei Ba, Jamie~Ryan Kiros, and Geoffrey~E. Hinton.
\newblock {Layer Normalization}.
\newblock Technical report, 2016.
\newblock URL \url{https://arxiv.org/abs/1607.06450}.

\bibitem[Bagdanov et~al.(2011)Bagdanov, Masi, and
  Del~Bimbo]{bagdanov2011florence}
Andrew~D. Bagdanov, Iacopo Masi, and Alberto Del~Bimbo.
\newblock The florence 2d/3d hybrid face datset.
\newblock In \emph{Proc. of ACM Multimedia Int.’l Workshop on Multimedia
  access to 3D Human Objects (MA3HO’11)}. ACM Press, December 2011.

\bibitem[Besl and McKay(1992)]{besl1992method}
Paul~J Besl and Neil~D McKay.
\newblock Method for registration of 3-d shapes.
\newblock In \emph{Sensor fusion IV: control paradigms and data structures},
  1992.

\bibitem[Bogo et~al.(2016)Bogo, Kanazawa, Lassner, Gehler, Romero, and
  Black]{bogo16keep}
F.~Bogo, A.~Kanazawa, C.~Lassner, P.~Gehler, J.~Romero, and M.~J. Black.
\newblock Keep it {SMPL}: Automatic estimation of {3D} human pose and shape
  from a single image.
\newblock In \emph{Proc. {ECCV}}, 2016.

\bibitem[Bregler et~al.(2000)Bregler, Hertzmann, and
  Biermann]{bregler2000recovering}
Christoph Bregler, Aaron Hertzmann, and Henning Biermann.
\newblock Recovering non-rigid {3D} shape from image streams.
\newblock In \emph{Proc. {CVPR}}, 2000.

\bibitem[Carreira et~al.(2015)Carreira, Kar, Tulsiani, and
  Malik]{carreira2015virtual}
Joao Carreira, Abhishek Kar, Shubham Tulsiani, and Jitendra Malik.
\newblock Virtual view networks for object reconstruction.
\newblock In \emph{Proc. {CVPR}}, 2015.

\bibitem[Cashman and Fitzgibbon(2013)]{cashman2013shape}
Thomas~J Cashman and Andrew~W Fitzgibbon.
\newblock What shape are dolphins? building 3d morphable models from 2d images.
\newblock \emph{{PAMI}}, 35\penalty0 (1):\penalty0 232--244, 2013.

\bibitem[Charbonnier et~al.(1997)Charbonnier, Blanc-f{\'e}raud, Aubert, and
  Barlaud]{Charbonnier97}
Pierre Charbonnier, Laure Blanc-f{\'e}raud, Gilles Aubert, and Michel Barlaud.
\newblock Deterministic edge-preserving regularization in computed imaging.
\newblock \emph{IEEE Trans. Image Processing}, 6:\penalty0 298--311, 1997.

\bibitem[Chen et~al.(2019)Chen, Gao, Ling, Smith, Lehtinen, Jacobson, and
  Fidler]{Chen2019}
Wenzheng Chen, Jun Gao, Huan Ling, Edward~J. Smith, Jaakko Lehtinen, Alec
  Jacobson, and Sanja Fidler.
\newblock {Learning to Predict 3D Objects with an Interpolation-based
  Differentiable Renderer}.
\newblock In \emph{Proc. {NeurIPS}}, 2019.
\newblock URL \url{http://arxiv.org/abs/1908.01210}.

\bibitem[Dai et~al.(2014)Dai, Li, and He]{dai2014simple}
Yuchao Dai, Hongdong Li, and Mingyi He.
\newblock A simple prior-free method for non-rigid structure-from-motion
  factorization.
\newblock \emph{International Journal of Computer Vision}, 107\penalty0
  (2):\penalty0 101--122, 2014.

\bibitem[Fragkiadaki et~al.(2014)Fragkiadaki, Salas, Arbelaez, and
  Malik]{fragkiadaki2014grouping}
Katerina Fragkiadaki, Marta Salas, Pablo Arbelaez, and Jitendra Malik.
\newblock Grouping-based low-rank trajectory completion and {3D}
  reconstruction.
\newblock In \emph{Proc. {NIPS}}, 2014.

\bibitem[Godard et~al.(2017)Godard, Aodha, and Brostow]{godard17unsupervised}
C.~Godard, O.~M. Aodha, and G.~J. Brostow.
\newblock Unsupervised monocular depth estimation with left-right consistency.
\newblock In \emph{Proc. {CVPR}}, 2017.

\bibitem[Gotardo and Martinez(2011)]{gotardo2011non}
Paulo~FU Gotardo and Aleix~M Martinez.
\newblock Non-rigid structure from motion with complementary rank-3 spaces.
\newblock In \emph{Proc. {CVPR}}, 2011.

\bibitem[Groueix et~al.(2018)Groueix, Fisher, Kim, Russell, and
  Aubry]{Groueix2018}
Thibault Groueix, Matthew Fisher, Vladimir~G. Kim, Bryan~C. Russell, and
  Mathieu Aubry.
\newblock {AtlasNet: A
  P$\backslash${\^{}}{\{}a{\}}pier-Mach$\backslash$'{\{}e{\}} Approach to
  Learning 3D Surface Generation}.
\newblock In \emph{Proc. {CVPR}}, pages 216--224, 2018.
\newblock \doi{10.1109/CVPR.2018.00030}.
\newblock URL \url{https://arxiv.org/abs/1802.05384}.

\bibitem[Guan et~al.(2009)Guan, Weiss, Balan, and Black]{guan09estimating}
P.~Guan, A.~Weiss, A.~O. Balan, and M.~J. Black.
\newblock Estimating human shape and pose from a single image.
\newblock In \emph{Proc. {ICCV}}, 2009.

\bibitem[G{\"u}ler et~al.(2018)G{\"u}ler, Neverova, and
  Kokkinos]{guler18densepose:}
A.~G{\"u}ler, N.~Neverova, and I.~Kokkinos.
\newblock {DensePose}: Dense human pose estimation in the wild.
\newblock In \emph{Proc. {CVPR}}, 2018.

\bibitem[Hariharan et~al.(2015)Hariharan, Arbel{\'{a}}ez, Girshick, and
  Malik]{Hariharan2015}
Bharath Hariharan, Pablo Arbel{\'{a}}ez, Ross Girshick, and Jitendra Malik.
\newblock {Hypercolumns for object segmentation and fine-grained localization}.
\newblock In \emph{Proc. {CVPR}}, 2015.

\bibitem[He et~al.(2016)He, Zhang, Ren, and Sun]{he2016deep}
Kaiming He, Xiangyu Zhang, Shaoqing Ren, and Jian Sun.
\newblock Deep residual learning for image recognition.
\newblock In \emph{Proc. {CVPR}}, 2016.

\bibitem[He et~al.(2017)He, Gkioxari, Doll{\'a}r, and Girshick]{he2017mask}
Kaiming He, Georgia Gkioxari, Piotr Doll{\'a}r, and Ross Girshick.
\newblock Mask r-cnn.
\newblock In \emph{Proc. {ICCV}}, 2017.

\bibitem[Henderson et~al.(2020)Henderson, Tsiminaki, and
  Lampert]{Henderson2020}
Paul Henderson, Vagia Tsiminaki, and Christoph~H. Lampert.
\newblock {Leveraging 2D Data to Learn Textured 3D Mesh Generation}.
\newblock In \emph{Proc. {CVPR}}, pages 7495--7504, 2020.
\newblock \doi{10.1109/cvpr42600.2020.00752}.

\bibitem[Huang et~al.(2017)Huang, Bogo, Lassner, Kanazawa, Gehler, Romero,
  Akhter, and Black]{huang17towards}
Yinghao Huang, Federica Bogo, Christoph Lassner, Angjoo Kanazawa, Peter~V.
  Gehler, Javier Romero, Ijaz Akhter, and Michael~J. Black.
\newblock Towards accurate marker-less human shape and pose estimation over
  time.
\newblock In \emph{Proc. {3DV}}, 2017.

\bibitem[Innmann et~al.(2019)Innmann, Kim, Gu, Niessner, Loop, Stamminger, and
  Kautz]{Innmann2019}
Matthias Innmann, Kihwan Kim, Jinwei Gu, Matthias Niessner, Charles Loop, Marc
  Stamminger, and Jan Kautz.
\newblock {NRMVS: Non-Rigid Multi-View Stereo}.
\newblock 2019.
\newblock URL \url{http://arxiv.org/abs/1901.03910}.

\bibitem[Joo et~al.(2018)Joo, Simon, and Sheikh]{joo18total}
Hanbyul Joo, Tomas Simon, and Yaser Sheikh.
\newblock Total capture: A {3D} deformation model for tracking faces, hands,
  and bodies.
\newblock In \emph{Proc. {CVPR}}, 2018.

\bibitem[Kanazawa et~al.(2018{\natexlab{a}})Kanazawa, Black, Jacobs, and
  Malik]{kanazawa18end-to-end}
Angjoo Kanazawa, Michael~J. Black, David~W. Jacobs, and Jitendra Malik.
\newblock End-to-end recovery of human shape and pose.
\newblock In \emph{Proc. {CVPR}}, 2018{\natexlab{a}}.

\bibitem[Kanazawa et~al.(2018{\natexlab{b}})Kanazawa, Tulsiani, Efros, and
  Malik]{kanazawa18learning}
Angjoo Kanazawa, Shubham Tulsiani, Alexei~A. Efros, and Jitendra Malik.
\newblock Learning category-specific mesh reconstruction from image
  collections.
\newblock In \emph{Proc. {ECCV}}, 2018{\natexlab{b}}.

\bibitem[Kar et~al.(2015)Kar, Tulsiani, Carreira, and Malik]{kar2015category}
Abhishek Kar, Shubham Tulsiani, Joao Carreira, and Jitendra Malik.
\newblock Category-specific object reconstruction from a single image.
\newblock In \emph{Proc. {CVPR}}, 2015.

\bibitem[Ke~Sun and Wang(2019)]{sun2019deep}
Dong~Liu Ke~Sun, Bin~Xiao and Jingdong Wang.
\newblock Deep high-resolution representation learning for human pose
  estimation.
\newblock In \emph{Proc. {CVPR}}, 2019.

\bibitem[Khot et~al.(2019)Khot, Agrawal, Tulsiani, Mertz, Lucey, and
  Hebert]{Khot2019}
Tejas Khot, Shubham Agrawal, Shubham Tulsiani, Christoph Mertz, Simon Lucey,
  and Martial Hebert.
\newblock {Learning Unsupervised Multi-View Stereopsis via Robust Photometric
  Consistency}.
\newblock 2019.
\newblock URL \url{http://arxiv.org/abs/1905.02706}.

\bibitem[Kolotouros et~al.(2019)Kolotouros, Pavlakos, and
  Daniilidis]{kolotouros19convolutional}
Nikos Kolotouros, Georgios Pavlakos, and Kostas Daniilidis.
\newblock Convolutional mesh regression for single-image human shape
  reconstruction.
\newblock In \emph{Proc. {CVPR}}, 2019.

\bibitem[Kulkarni et~al.(2019)Kulkarni, Gupta, and
  Tulsiani]{kulkarni19canonical}
Nilesh Kulkarni, Abhinav Gupta, and Shubham Tulsiani.
\newblock Canonical surface mapping via geometric cycle consistency.
\newblock In \emph{Proc. {ICCV}}, 2019.

\bibitem[Kulkarni et~al.(2020)Kulkarni, Gupta, Fouhey, and
  Tulsiani]{Kulkarni2020}
Nilesh Kulkarni, Abhinav Gupta, David~F. Fouhey, and Shubham Tulsiani.
\newblock {Articulation-aware Canonical Surface Mapping}.
\newblock In \emph{Proc. {CVPR}}, 2020.
\newblock URL \url{http://arxiv.org/abs/2004.00614}.

\bibitem[Kumar et~al.(2017)Kumar, Dai, and Li]{kumar2017spatial}
Suryansh Kumar, Yuchao Dai, and Hongdong Li.
\newblock Spatial-temporal union of subspaces for multi-body non-rigid
  structure-from-motion.
\newblock \emph{Pattern Recognition Journal}, 2017.

\bibitem[Kumar et~al.(2018)Kumar, Cherian, Dai, and Li]{kumar2018scalable}
Suryansh Kumar, Anoop Cherian, Yuchao Dai, and Hongdong Li.
\newblock Scalable dense non-rigid structure from motion: A grassmannian
  perspective.
\newblock In \emph{Proc. {CVPR}}, 2018.

\bibitem[Lassner et~al.(2017)Lassner, Romero, Kiefel, Bogo, Black, and
  Gehler]{lassner17unite}
Christoph Lassner, Javier Romero, Martin Kiefel, Federica Bogo, Michael~J.
  Black, and Peter~V. Gehler.
\newblock Unite the people: Closing the loop between {3D} and {2D} human
  representations.
\newblock In \emph{Proc. {CVPR}}, 2017.

\bibitem[Liu et~al.(2015)Liu, Luo, Wang, and Tang]{liu2015faceattributes}
Ziwei Liu, Ping Luo, Xiaogang Wang, and Xiaoou Tang.
\newblock Deep learning face attributes in the wild.
\newblock In \emph{Proc. {ICCV}}, December 2015.

\bibitem[Loper et~al.(2015)Loper, Mahmood, Romero, Pons-Moll, and
  and]{loper15smpl}
M.~Loper, N.~Mahmood, J.~Romero, G.~Pons-Moll, and M.~J.~Black and.
\newblock {SMPL}: A skinned multi- person linear model.
\newblock \emph{{ACM} Trans. on Graphics}, 2015.

\bibitem[Mescheder et~al.(2019)Mescheder, Oechsle, Niemeyer, Nowozin, and
  Geiger]{Mescheder2019}
Lars Mescheder, Michael Oechsle, Michael Niemeyer, Sebastian Nowozin, and
  Andreas Geiger.
\newblock {Occupancy Networks: Learning 3D Reconstruction in Function Space}.
\newblock In \emph{Proc. {CVPR}}, 2019.

\bibitem[Novotny et~al.(2017)Novotny, Larlus, and Vedaldi]{novotny17learning}
David Novotny, Diane Larlus, and Andrea Vedaldi.
\newblock Learning 3d object categories by looking around them.
\newblock In \emph{Proc. {ICCV}}, 2017.

\bibitem[Novotny et~al.(2018)Novotny, Larlus, and Vedaldi]{novotny18capturing}
David Novotny, Diane Larlus, and Andrea Vedaldi.
\newblock Capturing the geometry of object categories from video supervision.
\newblock \emph{{PAMI}}, 2018.

\bibitem[Novotny et~al.(2019)Novotny, Ravi, Graham, Neverova, and
  Vedaldi]{novotny19c3dpo}
David Novotny, Nikhila Ravi, Benjamin Graham, Natalia Neverova, and Andrea
  Vedaldi.
\newblock {C3DPO}: Canonical 3d pose networks for non-rigid structure from
  motion.
\newblock In \emph{Proc. {ICCV}}, 2019.

\bibitem[Omran et~al.(2018)Omran, Lassner, Pons-Moll, Gehle, and
  Schiele]{omran18neural}
Mohamed Omran, Christoph Lassner, Gerard Pons-Moll, Peter~V. Gehle, and Bernt
  Schiele.
\newblock Neural body fitting: Unifying deep learning and model based human
  pose and shape estimation.
\newblock In \emph{{3DV}}, 2018.

\bibitem[Pavlakos et~al.(2018)Pavlakos, Zhu, Zhou, and
  Daniilidis]{pavlakos18learning}
Georgios Pavlakos, Luyang Zhu, Xiaowei Zhou, and Kostas Daniilidis.
\newblock Learning to estimate {3D} human pose and shape from a single color
  image.
\newblock In \emph{Proc. {CVPR}}, 2018.

\bibitem[Pavlakos et~al.(2019)Pavlakos, Choutas, Ghorbani, Bolkart, Osman,
  Tzionas, and Black]{pavlakos19expressive}
Georgios Pavlakos, Vasileios Choutas, Nima Ghorbani, Timo Bolkart, Ahmed A.~A.
  Osman, Dimitrios Tzionas, and Michael~J. Black.
\newblock Expressive body capture: {3D} hands, face, and body from a single
  image.
\newblock In \emph{Proc. {CVPR}}, 2019.

\bibitem[Paysan et~al.(2009)Paysan, Knothe, Amberg, Romdhani, and
  Vetter]{bfm09}
Pascal Paysan, Reinhard Knothe, Brian Amberg, Sami Romdhani, and Thomas Vetter.
\newblock A 3d face model for pose and illumination invariant face recognition.
\newblock In \emph{The IEEE International Conference on Advanced Video and
  Signal Based Surveillance}, 2009.

\bibitem[Roberts(1963)]{Roberts1963}
Lawrence~Gilman Roberts.
\newblock \emph{{The Perception of Three-Dimensional Solids}}.
\newblock PhD thesis, Massachusets Institute of Technology, 1963.
\newblock URL
  \url{https://dspace.mit.edu/bitstream/handle/1721.1/11589/33959125-MIT.pdf}.

\bibitem[Rogez et~al.(2018)Rogez, Weinzaepfel, and Schmid]{rogez18lcr-net}
Gr{\'{e}}gory Rogez, Philippe Weinzaepfel, and Cordelia Schmid.
\newblock {LCR-Net++}: Multi-person {2D} and {3D} pose detection in natural
  images.
\newblock \emph{{PAMI}}, 2018.

\bibitem[Schmidt et~al.(2017)Schmidt, Newcombe, and
  Fox]{schmidt17self-supervised}
Tanner Schmidt, Richard~A. Newcombe, and Dieter Fox.
\newblock Self-supervised visual descriptor learning for dense correspondence.
\newblock \emph{{IEEE} Robotics and Automation Letters}, 2\penalty0 (2), 2017.

\bibitem[Sedaghat and Brox(2015)]{nima2015unsupervised}
Nima Sedaghat and Tomas Brox.
\newblock Unsupervised generation of a viewpoint annotated car dataset from
  videos.
\newblock In \emph{Proc. {ICCV}}, 2015.

\bibitem[Sigal et~al.(2008)Sigal, Balan, and Black]{sigal08combined}
Leonid Sigal, Alexandru Balan, and Michael~J. Black.
\newblock Combined discriminative and generative articulated pose and non-rigid
  shape estimation.
\newblock In \emph{Proc. {NIPS}}. 2008.

\bibitem[Simonyan and Zisserman(2015)]{Simonyan2015}
Karen Simonyan and Andrew Zisserman.
\newblock Very deep convolutional networks for large-scale image recognition.
\newblock In \emph{International Conference on Learning Representations}, 2015.

\bibitem[Sun et~al.(2019)Sun, Zhao, Jiang, Cheng, Xiao, Liu, Mu, Wang, Liu, and
  Wang]{Sun2019}
Ke~Sun, Yang Zhao, Borui Jiang, Tianheng Cheng, Bin Xiao, Dong Liu, Yadong Mu,
  Xinggang Wang, Wenyu Liu, and Jingdong Wang.
\newblock {High-Resolution Representations for Labeling Pixels and Regions}.
\newblock 2019.
\newblock URL \url{http://arxiv.org/abs/1904.04514}.

\bibitem[Tan et~al.(2017)Tan, Budvytis, and Cipolla]{tan17indirect}
V.~Tan, I.~Budvytis, and R.~Cipolla.
\newblock Indirect deep structured learning for {3D} human body shape and pose
  prediction.
\newblock In \emph{Proc. {BMVC}}, 2017.

\bibitem[Tatarchenko et~al.(2019)Tatarchenko, Richter, Ranftl, Li, Koltun, and
  Brox]{Tatarchenko2019}
Maxim Tatarchenko, Stephan~R. Richter, Ren{\'{e}} Ranftl, Zhuwen Li, Vladlen
  Koltun, and Thomas Brox.
\newblock {What Do Single-view 3D Reconstruction Networks Learn?}
\newblock In \emph{Proc. {CVPR}}, 2019.

\bibitem[Thewlis et~al.(2017)Thewlis, Bilen, and
  Vedaldi]{thewlis17bunsupervised}
J.~Thewlis, H.~Bilen, and A.~Vedaldi.
\newblock Unsupervised object learning from dense invariant image labelling.
\newblock In \emph{Proc. {NIPS}}, 2017.

\bibitem[Thewlis et~al.(2019)Thewlis, Albanie, Bilen, and
  Vedaldi]{thewlis19unsupervised}
J.~Thewlis, S.~Albanie, H.~Bilen, and A.~Vedaldi.
\newblock Unsupervised learning of landmarks via vector exchange.
\newblock In \emph{Proc. {ICCV}}, 2019.

\bibitem[Torresani et~al.(2008)Torresani, Hertzmann, and
  Bregler]{torresani2008nonrigid}
Lorenzo Torresani, Aaron Hertzmann, and Chris Bregler.
\newblock Nonrigid structure-from-motion: Estimating shape and motion with
  hierarchical priors.
\newblock \emph{{PAMI}}, 30\penalty0 (5):\penalty0 878--892, 2008.

\bibitem[Tulsiani et~al.(2020)Tulsiani, Kulkarni, and Gupta]{Tulsiani2020}
Shubham Tulsiani, Nilesh Kulkarni, and Abhinav Gupta.
\newblock {Implicit Mesh Reconstruction from Unannotated Image Collections}.
\newblock 2020.
\newblock URL \url{http://arxiv.org/abs/2007.08504}.

\bibitem[Tung et~al.(2017)Tung, Tung, Yumer, and
  Fragkiadaki]{tung17self-supervised}
Hsiao-Yu~Fish Tung, Hsiao-Wei Tung, Ersin Yumer, and Katerina Fragkiadaki.
\newblock Self-supervised learning of motion capture.
\newblock In \emph{Proc. {NIPS}}, 2017.

\bibitem[Varol et~al.(2018)Varol, Ceylan, Russel, Yang, Yumer, Laptev, and
  Schmid]{varol18bodynet}
G.~Varol, D.~Ceylan, B.~Russel, J.~Yang, E.~Yumer, I.~Laptev, and C.~Schmid.
\newblock {BodyNet}: Volumetric inference of {3D} human body shapes.
\newblock In \emph{Proc. {ECCV}}, 2018.

\bibitem[Vicente et~al.(2014)Vicente, Carreira, Agapito, and
  Batista]{vicente2014reconstructing}
Sara Vicente, Joao Carreira, Lourdes Agapito, and Jorge Batista.
\newblock Reconstructing {PASCAL VOC}.
\newblock In \emph{Proc. {CVPR}}, 2014.

\bibitem[Wah et~al.(2011)Wah, Branson, Welinder, Perona, and
  Belongie]{WahCUB_200_2011}
C.~Wah, S.~Branson, P.~Welinder, P.~Perona, and S.~Belongie.
\newblock {The Caltech-UCSD Birds-200-2011 Dataset}.
\newblock Technical Report CNS-TR-2011-001, California Institute of Technology,
  2011.

\bibitem[Wang et~al.(2019)Wang, Sridhar, Huang, Valentin, Song, and
  Guibas]{Wang2019}
He~Wang, Srinath Sridhar, Jingwei Huang, Julien Valentin, Shuran Song, and
  Leonidas~J. Guibas.
\newblock {Normalized Object Coordinate Space for Category-Level 6D Object Pose
  and Size Estimation}.
\newblock In \emph{Proc. {CVPR}}, pages 2637--2646, 2019.

\bibitem[Wu et~al.(2020)Wu, Rupprecht, and Vedaldi]{Wu2020}
Shangzhe Wu, Christian Rupprecht, and Andrea Vedaldi.
\newblock {Unsupervised Learning of Probably Symmetric Deformable 3D Objects
  from Images in the Wild}.
\newblock In \emph{Proc. {CVPR}}, 2020.
\newblock URL \url{http://arxiv.org/abs/1911.11130}.

\bibitem[Xiang et~al.(2019)Xiang, Joo, and Sheikh]{xiang19monocular}
Donglai Xiang, Hanbyul Joo, and Yaser Sheikh.
\newblock Monocular total capture: Posing face, body, and hands in the wild.
\newblock In \emph{Proc. {CVPR}}, 2019.

\bibitem[Xiang et~al.(2014{\natexlab{a}})Xiang, Mottaghi, and
  Savarese]{xiang14beyond}
Y.~Xiang, R.~Mottaghi, and S.~Savarese.
\newblock Beyond pascal: A benchmark for 3d object detection in the wild.
\newblock \emph{Proc. {WACV}}, 2014{\natexlab{a}}.

\bibitem[Xiang et~al.(2014{\natexlab{b}})Xiang, Mottaghi, and
  Savarese]{xiang2014beyond}
Yu~Xiang, Roozbeh Mottaghi, and Silvio Savarese.
\newblock Beyond {PASCAL}: A benchmark for 3d object detection in the wild.
\newblock In \emph{WACV}, 2014{\natexlab{b}}.

\bibitem[Xiao et~al.(2004)Xiao, Jin-xiang, and Kanade]{xiao2004dense}
Jing Xiao, Chai Jin-xiang, and Takeo Kanade.
\newblock A closed-form solution to non-rigid shape and motion recovery.
\newblock In \emph{Proc. {ECCV}}, 2004.

\bibitem[Yao et~al.(2018)Yao, Luo, Li, Fang, and Quan]{Yao2018}
Yao Yao, Zixin Luo, Shiwei Li, Tian Fang, and Long Quan.
\newblock {MVSNet: Depth inference for unstructured multi-view stereo}.
\newblock In \emph{Proc. {ECCV}}, 2018.

\bibitem[Zanfir et~al.(2018)Zanfir, Marinoiu, and
  Sminchisescu]{zanfir18monocular}
A.~Zanfir, E.~Marinoiu, and C.~Sminchisescu.
\newblock Monocular {3D} pose and shape estimation of multiple people in
  natural scenes --- the importance of multiple scene constraints.
\newblock In \emph{Proc. {CVPR}}, 2018.

\bibitem[Zhang et~al.(2018)Zhang, Isola, Efros, Shechtman, and Wang]{Zhang2018}
Richard Zhang, Phillip Isola, Alexei~A. Efros, Eli Shechtman, and Oliver Wang.
\newblock {The Unreasonable Effectiveness of Deep Features as a Perceptual
  Metric}.
\newblock In \emph{Proc. {CVPR}}, 2018.

\bibitem[Zhou et~al.(2016{\natexlab{a}})Zhou, Zhu, Derpanis, and
  Daniilidis]{zhou2016sparseness}
Xiaowei Zhou, Menglong Zhu, Kosta Derpanis, and Kostas Daniilidis.
\newblock Sparseness meets deepness: 3d human pose estimation from monocular
  video.
\newblock In \emph{Proc. {CVPR}}, 2016{\natexlab{a}}.

\bibitem[Zhou et~al.(2016{\natexlab{b}})Zhou, Zhu, Leonardos, and
  Daniilidis]{zhou2016sparse}
Xiaowei Zhou, Menglong Zhu, Spyridon Leonardos, and Kostas Daniilidis.
\newblock Sparse representation for {3D} shape estimation: A convex relaxation
  approach.
\newblock \emph{{PAMI}}, 2016{\natexlab{b}}.

\bibitem[Zhou et~al.(2019)Zhou, Barnes, Lu, Yang, and Li]{Zhou2019}
Yi~Zhou, Connelly Barnes, Jingwan Lu, Jimei Yang, and Hao Li.
\newblock {On the Continuity of Rotation Representations in Neural Networks}.
\newblock In \emph{Proc. {CVPR}}, 2019.
\newblock URL \url{http://arxiv.org/abs/1812.07035}.

\bibitem[Zhu et~al.(2014)Zhu, Huang, De~La~Torre, and Lucey]{zhu2014complex}
Yingying Zhu, Dong Huang, Fernando De~La~Torre, and Simon Lucey.
\newblock Complex non-rigid motion {3D} reconstruction by union of subspaces.
\newblock In \emph{Proc. {CVPR}}, 2014.

\end{thebibliography}
